\documentclass{article}
\usepackage{iclr2027_conference,times}
\iclrpreprint 
\usepackage{amsmath,amssymb,amsthm,graphicx,booktabs,array}
\usepackage{algorithm,algpseudocode}
\usepackage{hyperref}
\definecolor{airforceblue}{rgb}{0.36, 0.54, 0.66}
\definecolor{iceberg}{rgb}{0.44, 0.65, 0.82}
\definecolor{glaucous}{rgb}{0.38, 0.51, 0.71}
\definecolor{turkishrose}{rgb}{0.71, 0.45, 0.51}
\definecolor{rosegold}{rgb}{0.72, 0.43, 0.47}
\definecolor{lavenderpurple}{rgb}{0.59, 0.48, 0.71}
\definecolor{figteal}{HTML}{1F6E8C}
\hypersetup{colorlinks=true, citecolor=figteal, urlcolor=figteal, linkcolor=rosegold}
\usepackage{url}
\usepackage[capitalize]{cleveref}
\usepackage{multirow}
\usepackage[table]{xcolor}
\usepackage{graphicx}
\usepackage{subcaption}
\usepackage[normalem]{ulem}
\usepackage{overpic}

\title{DualHeads: Reward-tilted few-step diffusion made simple}
\author{%
Yuanzhi Zhu\textsuperscript{1}\thanks{Equal contribution.} \quad Eleftherios Tsonis\textsuperscript{1}\footnotemark[1] \quad Simo Alami Chehboune\textsuperscript{1} \\
\bf Xi Wang\textsuperscript{1} \quad St\'ephane Lathuili\`ere\textsuperscript{2} \quad Vicky Kalogeiton\textsuperscript{1} \\[0.6ex]
\normalfont \textsuperscript{1}LIX, \'Ecole Polytechnique, CNRS, IPP \quad \textsuperscript{2}Inria at Univ.\ Grenoble Alpes, CNRS, LJK}
\newcommand{\E}{\mathbb{E}}

\newcommand{\method}{DualHeads}
\begin{document}
\maketitle
\begin{figure}[!h]
\centering
\includegraphics[width=0.86\linewidth]{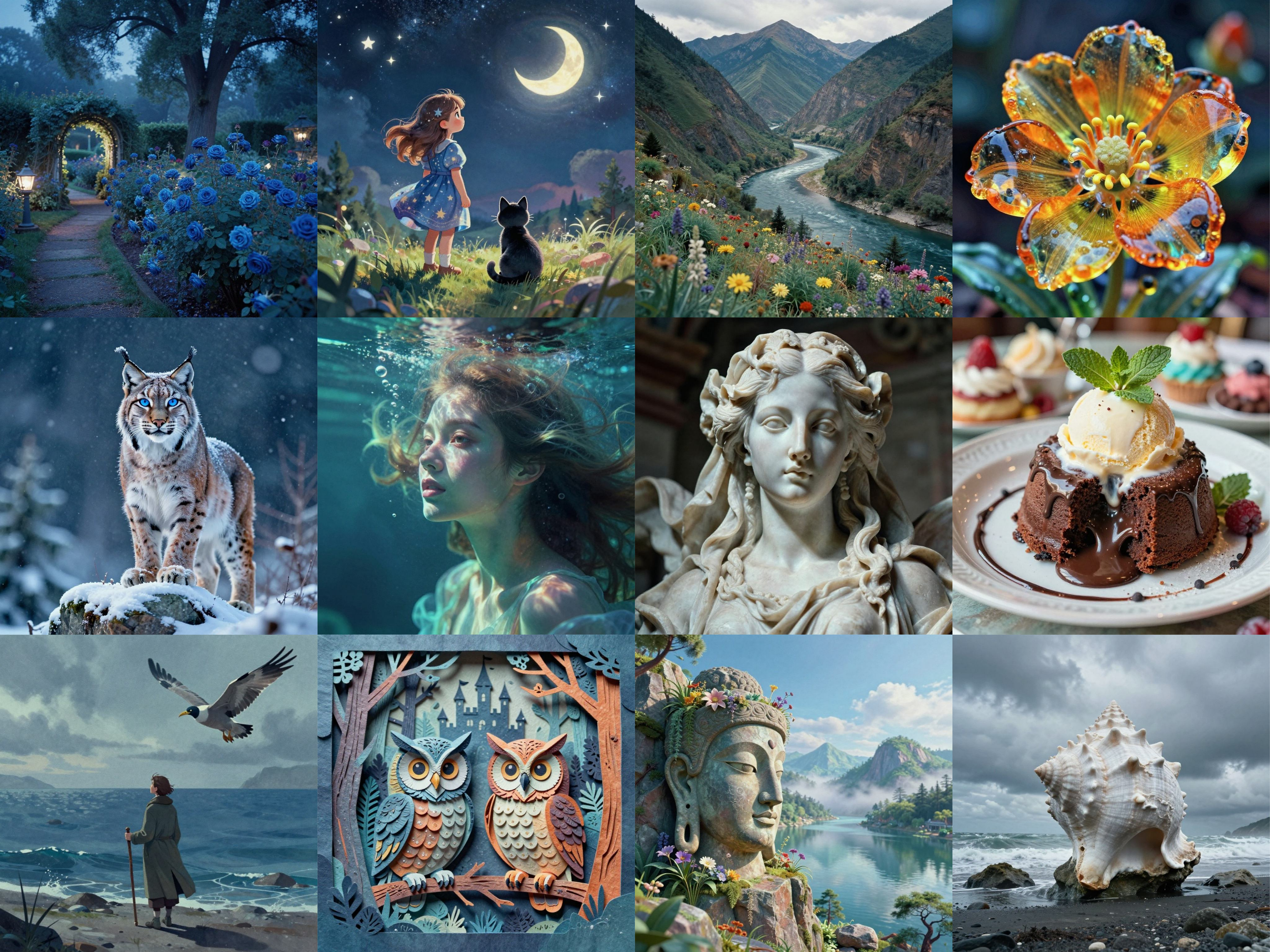}
\caption{\textbf{Four-step samples from \method{}.} Z-Image-Turbo~\citep{jiang2026distribution} fine-tuned with \method{} using the HPSv3 reward~\citep{ma2025hpsv3}. 
}
\label{fig:teaser}
\end{figure}

\begin{abstract}
Distilled diffusion models generate high-quality images in a few sampling steps, but aligning them with task requirements and human preferences is difficult. Methods developed for multi-step diffusion models rely on differentiable rewards, or on score fields and stochastic trajectories that distilled generators may lack. We ask whether a few-step generator can be adapted using only external reward evaluations and reference samples, and show that two binary classifiers suffice. Our method, \method{}, implements them as lightweight heads on a frozen diffusion backbone. The reward head learns which samples of the same prompt score higher, the reference head separates reference images from generated ones, and the generator maximizes a weighted sum of their logits. When the heads are optimal, this objective is KL-regularized maximization of the learned reward, and the ratio of head weights sets the regularization strength. Empirically, with both rule-based and model-based rewards, a few-step \method{} generator reaches task rewards close to those of 40-step reinforcement learning methods while preserving image quality, at no extra inference cost. We hope that this simple and elegant dual-head formulation will help advance few-step reward-tilted generation.
\end{abstract}

\section{Introduction}
\label{sec:introduction}

Diffusion distillation~\citep{song2023consistency,yin2024one,sauer2024fast,zhou2024score} turns powerful diffusion models~\citep{song2020score,ho2020denoising,rombach2022high,liu2022flow,lipman2022flow} into few-step generators that produce high-quality samples in a handful of network evaluations, bringing applications such as real-time interactive video generation within reach~\citep{yin2025slow,huang2026self}.

Speed alone, however, does not make a sample correct: a distilled student inherits the teacher's failures, and may still misspell the prompted text, miscount objects, or ignore the requested style. Such failures are readily caught by external evaluators, such as text recognizers, object detectors, or preference models. Using these external evaluators as rewards to post-train a generator is now standard practice for diffusion models~\citep{lee2023aligning,xu2023imagereward,prabhudesai2024video}.

Existing methods fall into two families, each typically addressing only one side of this setting. First, reinforcement learning (RL) with black-box rewards has succeeded for multi-step diffusion models, through policy gradients that reinforce high-reward denoising steps~\citep{black2024training,fan2023dpok,liu2025flow,xue2025dancegrpo} or forward-process objectives~\citep{zheng2025diffusionnft,xue2025advantage,choi2026scaling,lee2026reward}. These, however, rely on what distillation removes: policy gradients
estimate updates from stochastic multi-step rollouts, which become high-variance with only a few steps, and forward-process objectives need a score or velocity field that a distilled generator need not preserve.
Second, methods designed for few-step generators pay for the step constraint elsewhere: they backpropagate through a differentiable reward~\citep{luo2024diff} which may not always be feasible, %
match the generator to the original multi-step model (the teacher) using its score and a second diffusion model fitted to the generator's samples~\citep[e.g.,][]{luo2026tdm,jiang2026distribution,huang2026reinforcing}, or adapt multi-step RL specifically to flow-map generators such as consistency models and MeanFlow~\citep{li2026flow,huang2026meanflownft}.
Methods from both families rely on something beyond samples and black-box scores: trajectory likelihoods, a score field, reward gradients, or the teacher's score. 

In practice, however, the only thing any few-step generator is guaranteed to provide is samples. This leads to our central question: \emph{given only its own samples, black-box reward evaluations, and examples of high-quality images {from a reference model}, can we tilt a few-step generator toward higher reward regions while keeping it close to those images?} 

We show that this can be done with 
only one backbone network and two binary classifiers. We begin with the standard reference-regularized {RL} objective \citep{levine2018reinforcement},
which maximizes the expected reward while penalizing the reverse KL divergence to the reference distribution to prevent reward hacking. Its optimum is the reference distribution reweighted by the exponentiated reward. 
The two terms are, however, hard to optimize directly in our setting, since we treat the reward as a black box and the KL term depends on the density ratio between the generator and the reference. Our key observation is that both can be replaced by quantities learned through classification.

This leads to \method{}, a simple framework consisting of two {identical} lightweight heads on a shared frozen diffusion backbone. 
A \emph{reward head} learns which samples are better from their black-box rewards: for multiple generations of the same prompt, rewards are normalized within the group and converted into soft optimality labels. 
A \emph{reference head} distinguishes noised reference images from noised generator predictions.
Both heads are trained with the same binary cross-entropy objective and updated online as the generator %
updates. 
To update the generator, we evaluate the two heads on a noised generator prediction and maximize a weighted sum of their logits. Gradients pass through both heads and a single generator forward pass; the sampled trajectory and the head parameters remain fixed.
The heads are discarded at inference.  %
We note that at the optimum of the reference head, maximizing this weighted sum is equivalent to the reference-regularized objective from above, with the reward-head logit in place of the reward and the ratio of head weights setting the regularization strength.

We evaluate \method{} at a four-step setup on complementary modern reward functions, including visual text rendering (OCR), compositional generation (GenEval), and human preference (HPS, PickScore, etc.). 
{Compared with existing methods, \method{} achieves on-par performance in the few-step regime while preserving generation quality.} {It raises OCR accuracy from $0.59$ to $0.92$ and GenEval from $0.63$ to $0.90$ over the base model, and gives the best HPSv3 on Z-Image (\cref{fig:teaser}).}
 
Our contributions are:

{\setlength{\leftmargini}{1.5em}%
\begin{itemize}
    \item {We introduce} {\method{}, a simple framework that adapts few-step generators from reward evaluations and reference samples alone, using two classifier heads on a frozen backbone.}
    
    \item 
    {We show that, with optimal heads, the weighted-logit generator objective is a reverse-KL-regularized reward objective whose temperature is set by the head-weight ratio.}

    \item {We provide} 
    {empirical results with both rule-based and model-based rewards, showing that a four-step student reaches task rewards close to 40-step RL methods while preserving quality.}
    
\end{itemize}}

\section{Related Work 
}
\label{sec:related}

\paragraph{Step distillation of diffusion models.}
Progressive and consistency distillation~\citep{salimans2022progressive,song2023consistency} turn a pretrained diffusion model into a few-step generator that covers the sampling trajectory in a small number of large jumps. Latent consistency models~\citep{luo2023latent} extend this to text-to-image latent diffusion.
Distribution matching distillation (DMD)~\citep{yin2024one} instead matches the student's output distribution to the teacher's by following the gradient from a score difference, which requires an auxiliary network that tracks the student's score online. DMD2~\citep{yin2024improved} stabilizes this estimator and adds an adversarial term on real data, and TDM~\citep{luo2025learning} extends the matching to every step of the sampling trajectory.
Parallel work relies on adversarial supervision. ADD~\citep{sauer2024adversarial} distills with a discriminator on pretrained image features, and LADD~\citep{sauer2024fast} moves the discriminator onto frozen features of the teacher diffusion model itself.
\method{} is built on top of a few-step distilled diffusion model. Our reference head follows LADD's design of heads on frozen diffusion features.

\paragraph{Reward optimization of diffusion models.}
Methods for optimizing a reward differ mainly in how the reward signal reaches the generator. Differentiable rewards can be backpropagated through directly~\citep{xu2023imagereward,clark2024directly,prabhudesai2023aligning}, but black-box rewards cannot. Reinforcement learning (RL) removes this requirement: DDPO~\citep{black2024training} and DPOK~\citep{fan2023dpok} use policy gradients, and Flow-GRPO~\citep{liu2025flow} and DanceGRPO~\citep{xue2025dancegrpo} adapt group-relative policy optimization~\citep{shao2024deepseekmath} to flow models.
These methods need transition likelihoods along stochastic multi-step rollouts, which recent work avoids by training on the forward process.
DiffusionNFT~\citep{zheng2025diffusionnft} contrasts positive and negative samples under clipped group-relative optimality labels, CRD~\citep{zhu2026diffusion} matches prompt-centered rewards to cancel an intractable normalizer, and AWM~\citep{xue2025advantage}, RAM~\citep{bergmeister2026reinforce}, RVM~\citep{choi2026scaling}, and DGPO~\citep{luo2025reinforcing} use related advantage- or reward-weighted targets. In all of them, the reward serves only as a sample weight or a regression target, and no explicit reward model is learned.
\method{} uses labels of the same form as
DiffusionNFT, but fits them with an explicit classifier and guides the student
through that classifier's input gradient.

\paragraph{Reward optimization coupled with distillation.}
Other methods optimize the reward inside a distillation pipeline and differ mainly in what keeps the generator close to good images while the reward is optimized.
Most rely on a teacher. Some combine a reward objective with distribution matching~\citep{jiang2026distribution,chen2026flash,dong2026guiding,wang2026advdmd,fan2026r}; for example, RTDMD~\citep{huang2026reinforcing} minimizes a KL divergence to a reward-tilted teacher.
Closest to our setting, TDM-R1~\citep{luo2026tdm} post-trains a few-step student with non-differentiable rewards by learning a diffusion-model surrogate reward, and regularizes it with trajectory distribution matching, which requires the teacher and an online fake score.
Flow-OPD~\citep{fang2026flow} and DiffusionOPD~\citep{li2026diffusionopd} first train task-specific teachers with RL and then distill them into one student, without reducing the step count. REST~\citep{li2026rl} co-trains an eight-step student with an evolving teacher.
Others do without a teacher. MeanFlowNFT~\citep{huang2026meanflownft} applies forward-process RL to a distilled four-step generator, anchored to a frozen EMA of itself, and Self-OPD~\citep{zhang2026self} scores branched student rollouts against the student's own deterministic trajectory.
Reward Lightning~\citep{cheng2026reward} is closest in architecture: it jointly distills a video teacher into a few-step generator with discriminator and reward heads on one frozen backbone, but its reward head is pretrained on preference pairs and kept frozen.
\method{} instead post-trains a distilled generator and refits its reward head online, on pointwise group-relative labels from black-box rewards, while its reference head keeps the generator close to the reference pool.

\newcommand{\bce}{\operatorname{BCE}}
\newcommand{\logit}{\operatorname{logit}}
\newcommand{\xhat}{\hat{x}}

\begin{figure}[!t]
    \centering
    \vspace{-15pt}
\def\figW{0.95}\def\figS{1.1}%
\newcommand{\figsz}[2]{\scalebox{\fpeval{#1*\figW*\linewidth/(1151*10pt)*\figS}}{#2}}%
\def\figimgS{19.6}\def\figimgY{7.92}%
\begin{overpic}[width=\figW\linewidth]{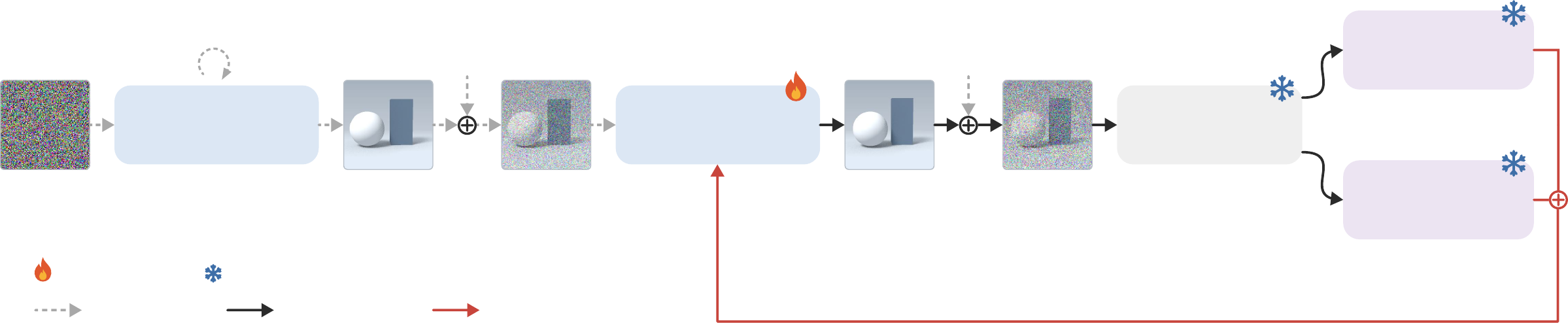}%
\put(2.92,\figimgY){\makebox[0pt]{\figsz{\figimgS}{$x_0\sim\mathcal{N}(0,I)$}}}
\put(10.18,16.32){\makebox(0,0){\figsz{13.4}{$N$ {steps}}}}
\put(13.85,13.51){\makebox(0,0){\figsz{16.9}{{Generator}}}}
\put(13.83,11.50){\makebox(0,0){\figsz{17.8}{$f_\theta$}}}
\put(24.71,\figimgY){\makebox[0pt]{\figsz{\figimgS}{$x_1$}}}
\put(29.80,16.86){\makebox(0,0){\figsz{16}{$\epsilon'$}}}
\put(35.12,\figimgY){\makebox[0pt]{\figsz{\figimgS}{$x_{t_i}$}}}
\put(45.82,13.51){\makebox(0,0){\figsz{16.9}{{Generator}}}}
\put(45.80,11.56){\makebox(0,0){\figsz{17.8}{$f_\theta$}}}
\put(57.03,\figimgY){\makebox[0pt]{\figsz{\figimgS}{$\hat{x}_1$}}}
\put(61.79,16.66){\makebox(0,0){\figsz{16}{$\epsilon$}}}
\put(67.18,\figimgY){\makebox[0pt]{\figsz{\figimgS}{$x_t$}}}
\put(77.19,13.16){\makebox(0,0){\figsz{15.6}{{Feature Extractor}}}}
\put(77.24,11.32){\makebox(0,0){\figsz{16.9}{$F(x_t,c,t)$}}}
\put(91.79,18.19){\makebox(0,0){\figsz{15.2}{{Reference head}}}}
\put(91.79,16.03){\makebox(0,0){\figsz{16.9}{$D_\phi$}}}
\put(91.78,8.63){\makebox(0,0){\figsz{15.2}{{Reward head}}}}
\put(91.84,6.47){\makebox(0,0){\figsz{16.5}{$R_\psi$}}}
\put(101.99,9.59){\makebox(0,0){\rotatebox{-90}{\figsz{16}{$s=\lambda_D D_\phi+\lambda_R R_\psi$}}}}
\put(73.87,1.56){\makebox(0,0){\figsz{17.4}{$\nabla_\theta\mathcal{L}_f(\theta)=-\mathbb{E}[\nabla_\theta s]$}}}
\put(3.74,2.62){\figsz{14.3}{{Trainable}}}
\put(14.60,2.62){\figsz{14.3}{{Frozen}}}
\put(6.17,0.28){\figsz{14.3}{{No gradient}}}
\put(18.40,0.28){\figsz{14.3}{{Differentiable}}}
\put(31.50,0.28){\figsz{14.3}{{Gradient to $\theta$}}}
\end{overpic}
    \vspace{-2pt}
    \caption{
    {\textbf{Generator update of \method{}.}} The generator $f_\theta$ first rolls out $N$ steps from noise without gradient; the endpoint $x_1$ is re-noised to $x_{t_i}$, from which $f_\theta$ predicts $\hat{x}_1$ in one differentiable step. The prediction is noised to $x_t$ and sent to a frozen feature extractor $F$ followed by a reference head $D_\phi$ and a reward head $R_\psi$; the generator ascends the utility $s=\lambda_D D_\phi+\lambda_R R_\psi$ through $\hat{x}_1$, while the backbone and both heads stay fixed during its update. 
    }
    \label{fig:pipeline}
\end{figure}

\section{Method}
\label{sec:method}

\method{} adapts a few-step generator so that its samples score higher under an external reward while staying close to a reference distribution. For this, we use two binary classifiers on a shared frozen backbone: %
one learns \emph{reference fidelity}, the other \emph{reward optimality}. 
\cref{sec:setup} describes the few-step generation setting and the corresponding reward-tilted target.
\cref{sec:dualheads}  defines the two classifier objectives and \cref{sec:generator_objective} gives the single utility the generator maximizes. %

\subsection{Background}
\label{sec:setup}

\paragraph{Setup of few-step generator.}
Let $\theta$ denote the parameters of a neural network few-step generator $f_\theta$; $c$ the frozen prompt embeddings and $x$ an image sample.
We write time in the flow-matching convention, with data at $t = 1$. For a clean sample $x_1$
and noise $\epsilon \sim \mathcal{N}(0, I)$, the interpolant at time $t \in [0, 1]$ and the
generator's clean-data prediction are: 
\begin{equation}
  x_t \;=\; \mathcal{N}_t(x_1; \epsilon)= t\, x_1 + (1 - t)\, \epsilon,
  \qquad
  f_\theta(x_t, c, t) \;=\; x_t + (1 - t)\, v_\theta(x_t, c, t),
  \label{eq:noising}
\end{equation}
where $v_\theta$ is the velocity representation of the model output. 
Starting from Gaussian noise $x_{t_0} \sim \mathcal{N}(0,I)$, the $N$-step stochastic
sampler follows a timestep grid $0=t_0<t_1<\cdots<t_N=1$.
At each step, the model predicts the clean endpoint and re-noises it to the next timestep:
\begin{equation}
x_{t_{i+1}}
=
\mathcal{N}_{t_{i+1}} \!\left( f_\theta(x_{t_i},c,t_i); \epsilon_{i+1} \right), \qquad i=0,\ldots,N-1, \label{eq:rollout}
\end{equation}
where $\epsilon_1,\ldots,\epsilon_N \overset{\mathrm{i.i.d.}}{\sim}\mathcal{N}(0,I)$ are independent Gaussian noise.
The noising operator satisfies {$\mathcal{N}_1(x;\epsilon)=x$ },
so the final step adds no noise and returns $x_{t_N}=f_\theta(x_{t_{N-1}},c,t_{N-1})$.
We write $p_\theta(x| c)$ for the distribution of the sampler's output $x=x_{t_N}$.

\paragraph{Goal: a reward-tilted reference distribution.}
We are given a black-box image reward $r(x,c)\in[0,1]$ and prompt-paired reference samples from a conditional distribution $p_{\mathrm{ref}}(x| c)$.
Our goal is to train a few-step generator whose output distribution $p_\theta(x| c)$ achieves high expected reward while remaining close to the reference distribution.
For a fixed prompt $c$, a standard formulation for KL-regularized RL training is:
\begin{align}
\mathcal{J}_c(\theta)
&= \mathrm{KL}\!\left(p_\theta(\cdot| c)\,\middle\|\,p_{\mathrm{ref}}(\cdot| c)\right) - \frac{1}{\beta}\mathbb{E}_{x\sim p_\theta(\cdot| c)}\big[r(x,c)\big],
\label{eq:objective}
\\
p^*(x| c)
&= \frac{1}{Z(c)}p_{\mathrm{ref}}(x| c)\exp\!\left(\frac{r(x,c)}{\beta}\right),
\label{eq:target}
\end{align}
where $\beta>0$ controls the strength of reference regularization relative to reward, $p^*(x| c)$ is the optimal solution to \cref{eq:objective}, and $Z(c)$ is the normalizing constant.
Indeed, $\mathcal{J}_c(\theta)=\mathrm{KL}(p_\theta\|p^*)-\log Z(c)$, so minimizing $\mathcal{J}_c$ is equivalent to fitting the reward-tilted target in reverse KL.

\subsection{\method{}}
\label{sec:dualheads}

Since the KL term requires unavailable density ratios and the black-box reward can be non-differentiable, neither term in \cref{eq:objective} is directly tractable. {To address this, we propose to} learn a classifier surrogate for each: a reference head and a reward head on a shared frozen backbone.

\paragraph{Two heads on one frozen backbone.}
{A pretrained frozen network serves as a feature extractor $F(x_t,c,t)$ for two lightweight heads with identical architecture: }
The \emph{reference head} $D_\phi$ distinguishes reference images $x_{\mathrm{ref}}\sim p_{\mathrm{ref}}$ from online generator 
predictions.
The \emph{reward head} $R_\psi$ instead predicts how good a sample is {based on some external evaluation score}. Its target is a soft label $y \in [0,1]$, computed from the reward of the rollout endpoint. 
Both heads are trained with the same binary cross-entropy loss, $\bce(\ell;y)=-y\,\log\sigma(\ell)\!-(1\!-\!y)\log\bigl(1\!-\!\sigma(\ell)\bigr)$. For brevity, we use $D_\phi(x_t,c,t)$ for $D_\phi(F(x_t,c,t),c,t)$ and likewise for $R_{\psi}$. 

\textbf{The generator prediction $\xhat_1$} is formed in two steps: a rollout endpoint $x_1$, sampled using \cref{eq:rollout}, is re-noised to a rollout grid timestep $t_i$, yielding $x_{t_i}=\mathcal{N}_{t_i}(x_1;\epsilon')$, and the generator is then applied in one differentiable step, yielding $\xhat_1=f_\theta(x_{t_i},c,t_i)$. 

\textbf{The reference head} uses hard targets: $1$ for noised reference images, $0$ for noised generator predictions, and its training objective is~\citep{goodfellow2014generative}:
\begin{equation}
\mathcal{L}_D(\phi) = \E_{t,\epsilon,x_{\mathrm{ref}}}\!\left[\bce\!\left(D_\phi(\mathcal{N}_t(x_{\mathrm{ref}};\epsilon),c,t);1\right)\right] + \E_{t,\epsilon,\xhat_1}\!\left[\bce\!\left(D_\phi(\mathcal{N}_t(\xhat_1;\epsilon),c,t);0\right)\right].
\label{eq:loss_disc}
\end{equation}

\textbf{The reward head} uses soft label targets 
$y$, which express reward preference \emph{relative to the other samples drawn for the same prompt}; we take $y$ as given here and show our construction of it at the end of this section.
The loss is identical in form to \cref{eq:loss_disc}\footnote{Pairwise preference data can be incorporated directly by assigning target $1$ to the winner and $0$ to the loser, yielding the same binary classification loss form as the reference head.}:
\begin{equation}
\mathcal{L}_R(\psi) = \E_{t,\epsilon,(x_1,y)}\!\left[\bce\!\left(R_\psi(\mathcal{N}_t(x_1;\epsilon),c,t);y\right)\right].
\label{eq:loss_reward}
\end{equation}
The label admits a probabilistic explanation: it defines a binary optimality variable $o\in\{0,1\}$ through $p(o=1\mid x_1,c)=y$, where $o=1$ denotes that $x_1$ is optimal among the samples for its prompt \citep{levine2018reinforcement}. 
In both objectives, $\epsilon\sim\mathcal{N}(0,I)$ noises the head's input at time $t\in[t_{\min},t_{\max}]$, drawn independently of the training example and of the $\epsilon'$ used to form $\xhat_1$.

The optimal heads under binary cross-entropy loss read:
\begin{equation}
\begin{aligned}
D^*(x_t,c,t) = \log\frac{p_{\mathrm{ref},t}(x_t\mid c)}{p_{\theta,t}^{\mathrm{pred}}(x_t\mid c)}, \quad R^*(x_t,c,t) = \log\frac{p(o=1\mid x_t,c,t)}{p(o=0\mid x_t,c,t)},
\end{aligned}
\label{eq:head_optima}
\end{equation}
where $p_{\mathrm{ref},t}$ and $p_{\theta,t}^{\mathrm{pred}}$ denote the noisy distributions of reference images and of generator predictions after noising by $\mathcal{N}_t$, and $p(o=1\mid x_t,c,t)$ is the probability that the rollout endpoint noised into $x_t$ is optimal.
For $D_\phi$, the target is $1$ on references and $0$ on predictions, so its conditional mean at $x_t$ is the share of the density at $x_t$ contributed by references; the corresponding log-odds is the log density ratio.
For $R_\psi$, the target is $y$ itself, so $\sigma(R^*)=\E[y\mid x_t,c,t]=p(o=1\mid x_t,c,t)$. 
In practice, both heads only approximate these targets, limited by the frozen backbone features and by optimization error.

\paragraph{Optimality reward labels.}

Following DiffusionNFT \citep{zheng2025diffusionnft}, we build $y$ from rewards normalized within each prompt group.
For each prompt $c$ we draw a group $\mathcal{G}=\{x^i\}_{i=1}^K$ of endpoints from the current generator, evaluate $r_i=r(x^i,c)$, and define
\begin{equation}
A_i = \frac{r_i-\bar r}{\operatorname{Std}(r)+\delta},
\qquad
y_i = \frac12+\frac12\,\operatorname{clip}\!\left(\frac{A_i}{C},-1,1\right),
\label{eq:labels}
\end{equation}
where $\bar r$ and $\operatorname{Std}(r)$ are the group mean and standard deviation, $\delta>0$ stabilizes the normalization, and $C{>}0$ controls the labels' sharpness.
For the $i$-th sample of $\mathcal{G}$, the optimality variable from above therefore satisfies $p(o_i=1\mid x^i,c,\mathcal{G})=y_i$. 
The labels average to $\tfrac12$ when no advantage is clipped.

\subsection{Generator objective}
\label{sec:generator_objective}

With both heads fixed, we train the generator to maximize their weighted logits:
\begin{equation}
\begin{aligned}
\mathcal{L}_f(\theta)
&= \mathbb{E}_{c,\xhat_1,t,\epsilon}\!\left[-s\!\left(\mathcal{N}_t(\xhat_1;\epsilon),c,t\right)\right],
\\
s(x_t,c,t)
&= \lambda_D D_\phi(x_t,c,t)+\lambda_R R_\psi(x_t,c,t),
\end{aligned}
\label{eq:utility}
\end{equation}
where $\lambda_D,\lambda_R\geq0$, $\epsilon\sim\mathcal{N}(0,I)$, and $t$ is sampled as in head training. 
The expectation averages over prompts, generator predictions, and noising randomness.
During each generator update, the head parameters and sampled rollout inputs remain fixed, while gradients propagate through the heads and frozen backbone to the prediction $\xhat_1=f_\theta(x_{t_i},c,t_i)$.

\paragraph{Reverse-KL interpretation.}
Given an optimal reference head, \cref{eq:utility} reduces to a standard KL-regularized reward objective.
Fix a prompt $c$, a noise level $t$ and the reward head.
Since $D^*$ in \cref{eq:head_optima} is a log density ratio, $\mathbb{E}_{p_{\theta,t}^{\mathrm{pred}}}[D^*]=-\mathrm{KL}(p_{\theta,t}^{\mathrm{pred}}\,\|\,p_{\mathrm{ref},t})$, and substituting it into \cref{eq:utility} gives
\begin{equation}
\begin{aligned}
\mathcal{L}_{f,c,t}^{*}(\theta)
&=
\lambda_D\,\mathrm{KL}\!\left(
p_{\theta,t}^{\mathrm{pred}}(\cdot\mid c)
\,\middle\|\,
p_{\mathrm{ref},t}(\cdot\mid c)
\right)
-\lambda_R\,\mathbb{E}_{x_t\sim p_{\theta,t}^{\mathrm{pred}}(\cdot\mid c)}
\!\left[R_\psi(x_t,c,t)\right].
\end{aligned}
\label{eq:generator_interpretation}
\end{equation}
The first reference term penalizes the reverse KL to the reference distribution, while the second reward term favors large reward-head logits.

\paragraph{Connection to reward tilting.}
For $\lambda_D,\lambda_R>0$, \cref{eq:generator_interpretation} is the standard KL-regularized objective \citep{schulman2017proximal,ziegler2019fine,ouyang2022training}, whose minimizer is
\begin{equation}
p_t^*(x_t\mid c)\propto p_{\mathrm{ref},t}(x_t\mid c)\exp\!\left(\frac{\lambda_R}{\lambda_D}R_\psi(x_t,c,t)\right),
\label{eq:fixed_point_target}
\end{equation} 
which has the same form as \cref{eq:target}, with $R_\psi$ in place of $r$ and $\lambda_D/\lambda_R$ in place of $\beta$.
The ratio of the head weights therefore sets the regularization strength.
Note that \cref{eq:fixed_point_target} holds for the noised predictions at each $t$, and that $R_\psi$ is fitted on samples from the current generator, so $p_t^*$ is a fixed point of training.
For unclipped labels close to $\tfrac12$, $R^*\approx\tfrac{2}{C}\,\mathbb{E}[A\mid x_t,c,t]$, which gives the equivalent temperature $\beta_{\mathrm{eq}}=C\lambda_D/(2\lambda_R)$ on the scale of the normalized advantage in \cref{eq:labels}.

\paragraph{Training and inference.}
Training alternates between classifier and generator updates at each iteration (\cref{alg:dualheads}). 
For each prompt in the batch, 
we sample $K$ few-step rollouts and evaluate their black-box rewards. 
The resulting rewards are used to calculate the soft label and then supervise the reward head. The same rollouts are re-noised to intermediate generator timesteps and passed once through $f_\theta$ to obtain predictions $\hat{x}_1$; 
the reference head is trained to distinguish these predictions from the prompt-paired reference images.
After updating both heads, we freeze them and update the generator through the weighted classifier logits (\cref{fig:pipeline}). 
After training, we discard both heads; hence, at inference, we only use the fine-tuned few-step generator, at no extra inference cost.

\begin{algorithm}[t]
\caption{\method{}}
\label{alg:dualheads}
\begin{algorithmic}[1]
\Require prompts with paired reference images; frozen backbone $F$; generator $f_\theta$; heads $D_\phi$, $R_\psi$; reward $r$; $\lambda_D$, $\lambda_R$, $C$, group size $K$
\State For each prompt $c$, run the sampler \cref{eq:rollout} $K$ times; score the endpoints $x_1$ with $r$
\State Form soft labels $y$ from group-relative advantages \Comment{\cref{eq:labels}}
\State Re-noise each $x_1$ to a grid time $t_i$; recompute $\xhat_1 = f_\theta(x_{t_i}, c, t_i)$
\State Update $\phi$ on $\mathcal{L}_D$ with $x_{\mathrm{ref}}$ against $\xhat_1$, and $\psi$ on $\mathcal{L}_R$ with $x_1$ and $y$ \Comment{\cref{eq:loss_disc,eq:loss_reward}}
\State Update $\theta$ by ascending $s\!\left(\mathcal{N}_t(\xhat_1;\epsilon),c,t\right)$ through $\xhat_1$, heads fixed \Comment{\cref{eq:utility}}
\end{algorithmic}
\end{algorithm}

\section{Experiments}
\label{sec:experiments}

\begin{table*}[t]
\vspace{-25pt}
\centering
\footnotesize
\setlength{\tabcolsep}{3pt}
\caption{Comparison on compositional image generation (\texttt{GenEval}), visual text rendering (\texttt{OCR}), and human preference metrics. Image quality and preference scores are computed on \texttt{DrawBench}~\citep{saharia2022photorealistic} prompts. Baseline numbers are from Flow-GRPO~\citep{liu2025flow} where reported, and from our reimplementation otherwise. Abbreviations: \texttt{Aes.} (Aesthetics), \texttt{ImgRwd} (ImageReward), \texttt{Pick} (PickScore), \texttt{CLIP} (CLIPScore), \texttt{HPS} (HPSv2.1). \texttt{NFE}: number of function evaluations. Few-step RL baselines are quoted from their papers; RTDMD reports only \texttt{GenEval} for its \texttt{GenEval}-only run. 
}
\vspace{-10pt}
\label{tab:main}
\resizebox{\textwidth}{!}{%
\begin{tabular}{@{}clcccccccc@{}}
\toprule
 & \textbf{Method} & \texttt{NFE} & \texttt{GenEval}$\uparrow$ & \texttt{OCR}$\uparrow$ & \texttt{Aes.}$\uparrow$ & \texttt{ImgRwd}$\uparrow$ & \texttt{Pick}$\uparrow$ & \texttt{CLIP}$\uparrow$ & \texttt{HPS}$\uparrow$ \\
  \cmidrule(lr){4-5} \cmidrule(lr){6-10} 
  & & & \multicolumn{2}{c}{{Task Metric}} & \multicolumn{5}{c}{{Held-out Metric}} \\ 
 \midrule
  & SD3.5-M (base) & 80 & 0.63 & 0.59 & 5.39 & 0.87 & 22.34 & 27.99 & 0.279 \\
  & SD3.5-M TDM~\citep{luo2025learning} & 4 & 0.62 & 0.51 & 5.58 & 0.86 & 22.14 & 27.18 & 0.290 \\
  \addlinespace
  \multicolumn{10}{@{}l}{\textit{Compositional image generation}} \\
  \cmidrule(r){1-2}
   & Flow-GRPO~\citep{liu2025flow} w/o KL & 80 & 0.95 & — & 4.93 & 0.44 & 21.16 & — & — \\
    & Flow-GRPO~\citep{liu2025flow} w/ KL & 80 & 0.95 & — & 5.25 & 1.03 & 22.37 & 29.25 & 0.274 \\
      & AWM~\citep{xue2025advantage} w/ CFG & 80 & 0.86 & — & 5.25 & 1.06 & 22.12 & 28.86 & 0.278 \\
     & AWM~\citep{xue2025advantage} w/o CFG & 40 & 0.91 & — & 5.10 & 0.39 & 21.76 & 26.99 & 0.235 \\
      & DiffusionNFT~\citep{zheng2025diffusionnft} & 40 & 0.92 & — & 5.30 & 0.63 & 22.07 & 27.25 & 0.253 \\ 
      & CRD~\citep{zhu2026diffusion} & 40 & {0.93} & — & {5.44} & {0.98} & {22.48} & {28.44} & {0.284} \\
     \multirow{-7}{*}{\rotatebox[origin=c]{90}{\scriptsize Multi-step RL}} & RSM~\citep{lee2026reward} & 40 & {0.97} & — & — & — & {22.28} & {31.60} & — \\
      \cmidrule(lr){1-10}
           & TDM-R1~\citep{luo2026tdm}  & 4 & 0.92 & — & 5.42 & 1.11 & 22.39 & --- & --- \\
           & RTDMD~\citep{huang2026reinforcing}  & 4
            & 0.94 & --- & --- & --- & --- & --- & --- \\
            & \color{gray} REST~\citep{li2026rl}$^{\ddagger}$ & \color{gray} 8 & \color{gray} 0.94 & \color{gray} --- & \color{gray} 5.67 & \color{gray} 1.05 & \color{gray} 22.79 & \color{gray} --- & \color{gray} 0.300 \\
           \rowcolor{gray!15} \cellcolor{white} & \textbf{\method{}} & 2 & {0.89} & — & {5.53} & {1.01} & {22.42} & {28.56} & {0.281} \\
           \rowcolor{gray!15} \cellcolor{white}\multirow{-5}{*}{\rotatebox[origin=c]{90}{\scriptsize Few-step RL}} & {\textbf{\method{}}} & 4 & {0.90} & — & {5.50} & {1.01} & {22.45} & {28.51} & {0.285} \\
           \addlinespace
           \multicolumn{10}{@{}l}{\textit{Visual text rendering}} \\
           \cmidrule(r){1-2}
            & Flow-GRPO~\citep{liu2025flow} w/o KL & 80 & — & 0.93 & 5.13 & 0.58 & 21.79 & — & — \\
             & Flow-GRPO~\citep{liu2025flow} w/ KL & 80 & — & 0.92 & 5.32 & 0.95 & 22.44 & 28.86 & 0.282 \\
               & AWM~\citep{xue2025advantage} w/ CFG & 80 & — & 0.96 & 5.35 & 0.96 & 22.36 & 28.70 & 0.283 \\
              & AWM~\citep{xue2025advantage} w/o CFG & 40 & — & 0.97 & 5.22 & $-$0.64 & 20.82 & 24.54 & 0.203 \\
               & DiffusionNFT~\citep{zheng2025diffusionnft} & 40 & — & 0.97 & 4.89 & $-$0.81 & 20.53 & 24.08 & 0.196 \\
               \multirow{-6}{*}{\rotatebox[origin=c]{90}{\scriptsize Multi-step RL}} & CRD~\citep{zhu2026diffusion} & 40 & — & 0.92 & 5.33 & 0.87 & 22.40 & 28.48 & 0.281 \\
               \cmidrule(lr){1-10}
               & TDM-R1~\citep{luo2026tdm} & 4 & — & 0.95 & 5.45 & 1.09 & 22.62 & — & — \\
               & DMSampler~\citep{liudistillation} & 4 & — & 0.83 & — & — & — & — & — \\
                & \color{gray} REST~\citep{li2026rl}$^{\ddagger}$ & \color{gray} 8 & \color{gray} — & \color{gray} 0.96 & \color{gray} 5.65 & \color{gray} 1.00 & \color{gray} 22.83 & \color{gray} — & \color{gray} 0.300 \\
               \rowcolor{gray!15} \cellcolor{white} & \textbf{\method{}} & 2 & — & {0.91} & {5.41} & {1.06} & {22.50} & {29.06} & {0.293} \\
               \rowcolor{gray!15} \cellcolor{white} & \textbf{\method{}} & 4 & — & {0.92} & {5.46} & {1.10} & {22.65} & {28.75} & {0.300} \\ 
               \rowcolor{gray!15} \cellcolor{white}\multirow{-6}{*}{\rotatebox[origin=c]{90}{\scriptsize Few-step RL}} & {\textbf{\method{}} + Nano Banana pool} & 4 & — & {0.92} & {5.62} & {1.08} & {22.59} & {26.86} & {0.298} \\
               \bottomrule
               \multicolumn{10}{@{}l}{$^{\ddagger}$REST is trained jointly on the task reward and \texttt{PickScore}.}
               \end{tabular}%
}
\vspace{-15pt}
\end{table*}

\subsection{Experimental setup}
\label{sec:protocol}

\paragraph{Models and rewards.}
We use SD3.5 Medium (SD3.5-M) and Z-Image as base models, trained at a resolution of $512^2$.
With SD3.5-M, we study two non-differentiable rewards, OCR and GenEval, trained separately, and one multi-reward setting, PickScore + CLIPScore + HPSv2.1. For Z-Image, we use HPSv3 as reward. 
The generators are initialized as few-step distilled diffusion models. 

\textbf{Prompt dataset.}
For OCR and GenEval, we use the corresponding training and test sets from Flow-GRPO.
For all other rewards, we train on Pick-a-Pic~\citep{kirstain2023pick} and evaluate on DrawBench~\citep{saharia2022photorealistic}.
Regardless of the training reward, we additionally evaluate every trained model on the OCR and GenEval test sets using their respective rewards.

\textbf{Reference pools.}
For OCR and GenEval, we generate reference pools with the base model using training-set prompts and 20 steps of sampling.
For OCR, we generate eight candidates for each of 5{,}000 prompts and retain the highest-scoring candidate for each prompt only if its OCR reward is at least $0.6$.
For GenEval, we construct a separate pool using 10{,}000 rows sampled from the GenEval training set.
For all other rewards, we use the existing Pick-a-Pic images as the reference pool.
Alternatively, an external model can generate the reference pool from the same training prompts{: we use Nano Banana, Google's proprietary image-generation model~\citep{google2026gemini31flashimage}}.

\textbf{Training and evaluation.}
Following prior work, we fine-tune the student generator using LoRA~\citep{hu2022lora} with rank $32$ and $\alpha = 64$.
Each epoch consists of 48 prompt groups, with group size $K {=} 24$ for the main results and $K {=} 8$ for all ablations.
{By default, the reward head is warmed up on samples from the reference distribution with scores from the reward, so that the generator does not receive noisy gradients early in training. This warm-up uses 1000 steps, a very small training budget compared with a full run (\cref{sec:exp-head}).}
Both training and evaluation use four sampling steps, unless stated otherwise, with the LCM sampler~\citep{luo2023latent} and no classifier-free guidance (CFG scale $1$).
We also evaluate the trained models using Aesthetic and ImageReward on the same evaluation set.
Details in \cref{app:implementation}.

\subsection{Task performance}
\label{sec:exp-task}

\paragraph{Text rendering.}
{With four steps, \method{} raises OCR accuracy from $0.59$ to $0.92$ ($0.91$ with two), on par with Flow-GRPO with KL and CRD (\cref{tab:main}). AWM without CFG and DiffusionNFT reach $0.97$ but lose image quality (negative ImageReward; HPS $\approx0.20$ vs.\ $0.279$ for the base model). \method{} instead improves {all quality and preference scores over the base model}, with the highest ImageReward ($1.10$) and HPS ($0.300$, tied with REST) among OCR-trained models. Among few-step methods, only TDM-R1 ($0.95$) and REST ($0.96$) score higher, using the teacher's score and 8 NFE with an extra PickScore reward, respectively.}

\begin{figure}[t]
\vspace{-35pt}
\centering
\begin{overpic}[width=\linewidth]{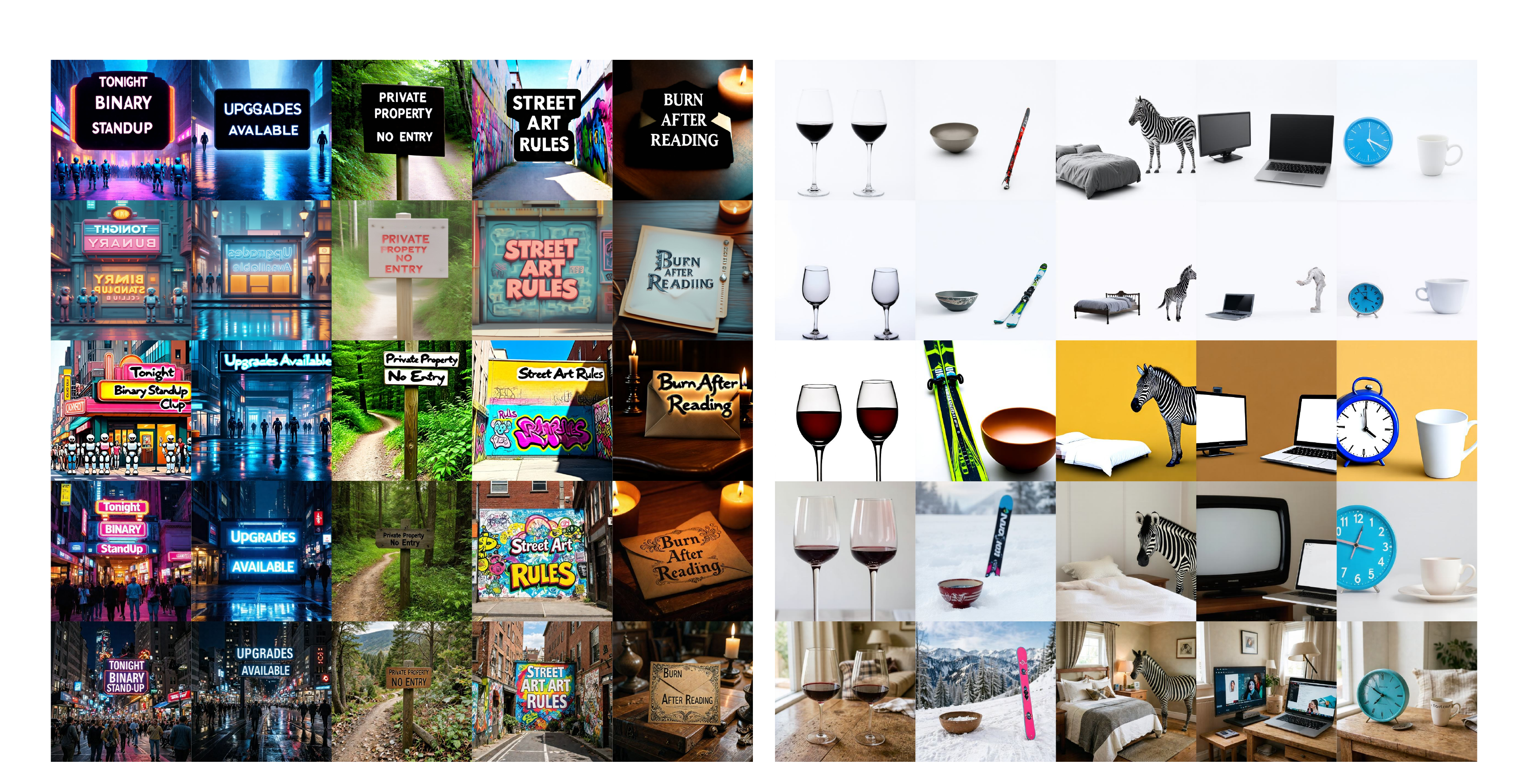}
\put(26.27,49.03){\makebox(0,0){\footnotesize OCR}}
\put(73.67,49.03){\makebox(0,0){\footnotesize GenEval}}
\put(2.09,42.81){\makebox(0,0){\rotatebox{90}{\fontsize{5}{6}\selectfont MeanFlowNFT}}}
\put(2.09,33.62){\makebox(0,0){\rotatebox{90}{\fontsize{5}{6}\selectfont RTDMD}}}
\put(2.09,24.43){\makebox(0,0){\rotatebox{90}{\fontsize{5}{6}\selectfont TDM-R1}}}
\put(2.09,15.24){\makebox(0,0){\rotatebox{90}{\fontsize{5}{6}\selectfont \method{}}}}
\put(2.09,6.05){\makebox(0,0){\rotatebox{90}{\fontsize{5}{6}\selectfont \method{} + NB}}}
\put(97.85,42.81){\makebox(0,0){\rotatebox{-90}{\fontsize{5}{6}\selectfont DiffusionNFT}}}
\put(97.85,33.62){\makebox(0,0){\rotatebox{-90}{\fontsize{5}{6}\selectfont RSM}}}
\put(97.85,24.43){\makebox(0,0){\rotatebox{-90}{\fontsize{5}{6}\selectfont TDM-R1}}}
\put(97.85,15.24){\makebox(0,0){\rotatebox{-90}{\fontsize{5}{6}\selectfont \method{}}}}
\put(97.85,6.05){\makebox(0,0){\rotatebox{-90}{\fontsize{5}{6}\selectfont \method{} + NB}}}
\end{overpic}
\vspace{-20pt}
\caption{\textbf{Qualitative comparison on OCR and GenEval test prompts.} \method{} + NB signifies training with the Nano Banana image pool.}
\label{fig:compare-verifiable}
\end{figure}

\paragraph{Compositional generation.}
{On GenEval, \method{} raises the overall score from $0.63$ to $0.90$, within the range of multi-step methods ($0.86$--$0.97$) and slightly below the few-step baselines ($0.92$--$0.94$), which again use the teacher's score or 8 NFE. Position ($0.24\to0.90$) and counting ($0.50\to0.91$) gain the most; attribute binding remains the weakest ($0.76$; \cref{tab:geneval} in \cref{app:geneval}). \method{} again improves all quality and preference scores over the base model, whereas Flow-GRPO without KL, AWM without CFG and DiffusionNFT lower ImageReward ($0.39$--$0.63$) and PickScore.}

\paragraph{Qualitative comparison.}
{In \cref{fig:compare-verifiable}, \method{} renders all five OCR texts correctly and blends them into the scene (neon signs, a trail sign, a mural, an envelope). MeanFlowNFT prints the text on flat black plates and misspells ``Upgrades Available'', RTDMD mirrors or misspells words (``Propety'', ``Readiing''), and TDM-R1 adds a word (``Clup''). On GenEval, the baselines draw isolated objects on plain backgrounds (RSM also misses the TV), while \method{} often places them in context, e.g., a ski in snow or a zebra beside a bed. The Nano Banana pool makes samples more photographic; it keeps OCR at $0.92$ but lowers GenEval to $0.73$ (\cref{sec:exp-head}).}

\paragraph{Model-based reward on image quality.}
{On Z-Image-Turbo with HPSv3 as the reward (\cref{tab:zimage}), four-step \method{} raises HPSv3 from $11.23$ to $13.93$, above TDM-R1 ($12.67$) and the 100-NFE base model ($9.93$). The gain carries over to held-out preference models, with the best PickScore ($23.03$) and HPSv2.1 ($0.314$) and a higher ImageReward ($0.96\to1.07$), while GenEval and OCR stay close to Z-Image-Turbo. Aesthetic and CLIPScore drop slightly, and TDM-R1 keeps a higher ImageReward and CLIPScore (samples in \cref{fig:compare-model-based}). With a PickScore + CLIPScore + HPSv2.1 reward on SD3.5-M, \method{} improves all five metrics over the base model but trails MeanFlowNFT and RTDMD on most (\cref{tab:multi_reward} in \cref{app:multi_reward}).}

\begin{table}[t]
\vspace{-10pt}
\centering
\footnotesize
\setlength{\tabcolsep}{3pt}
\caption{\textbf{Reinforcing Z-Image}~\citep{zimage2025} with \texttt{HPSv3}~\citep{ma2025hpsv3} as the training reward ($^{\ast}$). Z-Image-Turbo and TDM-R1~\citep{luo2026tdm} are evaluated at $1024^2$, except the entries marked $^{\dagger}$, which are reported in Table~3 of TDM-R1 (at $512^2$, with Z-Image-Turbo at 4 NFE); the \method{} rows are EMA weights of one run (reward weight 25) after 120 training iterations, sampled with two and four steps. \texttt{NFE}: number of function evaluations; \texttt{GenEval}: official overall accuracy; \texttt{OCR}: mean OCR score; \texttt{ImgRwd}: ImageReward; \texttt{CLIP}: CLIPScore. The best entry per column is in \textbf{bold}; shaded rows denote our method.}
\vspace{-5pt}
\label{tab:zimage}
\resizebox{\linewidth}{!}{%
\begin{tabular}{@{}lccccccccc@{}}
\toprule
& & \multicolumn{2}{c}{\textit{Verifiable metrics}} & \multicolumn{6}{c}{\textit{Model-based metrics}} \\
\cmidrule(lr){3-4} \cmidrule(l){5-10}
\textbf{Model} & \texttt{NFE} & \texttt{GenEval}$\uparrow$ & \texttt{OCR}$\uparrow$ & \texttt{HPSv3}$^{\ast}\uparrow$ & \texttt{Aes.}$\uparrow$ & \texttt{ImgRwd}$\uparrow$ & \texttt{Pick}$\uparrow$ & \texttt{CLIP}$\uparrow$ & \texttt{HPSv2.1}$\uparrow$ \\
\midrule
Z-Image (base;~\citealp{zimage2025}) & 100 & 0.66$^{\dagger}$ & 0.74$^{\dagger}$ & 9.93 & 5.36 & 0.92 & 22.39 & 28.47 & 0.285 \\
Z-Image-Turbo (DMDR;~\citealp{jiang2026distribution}) & 4 & 0.73$^{\dagger}$ & 0.78$^{\dagger}$ & 11.23 & \textbf{5.42} & 0.96 & 22.86 & 27.42 & 0.296 \\
TDM-R1~\citep{luo2026tdm} & 4 & \textbf{0.77}$^{\dagger}$ & \textbf{0.79}$^{\dagger}$ & 12.67 & 5.39 & \textbf{1.16} & 22.82 & \textbf{28.52} & 0.307 \\
\midrule
\rowcolor{gray!15} \method{}  & 2 & 0.76 & 0.64 & 12.91 & 5.27 & 1.10 & 22.77 & 27.31 & 0.307 \\
\rowcolor{gray!15} \method{}  & 4 & 0.74 & 0.75 & \textbf{13.93} & 5.36 & 1.07 & \textbf{23.03} & 27.12 & \textbf{0.314} \\
\bottomrule
\end{tabular}%
}
\end{table}

\subsection{Ablations}
\label{sec:exp-head}
\begin{figure*}[t]
  \centering
  \begin{subfigure}{0.5\textwidth}
    \includegraphics[width=\linewidth]{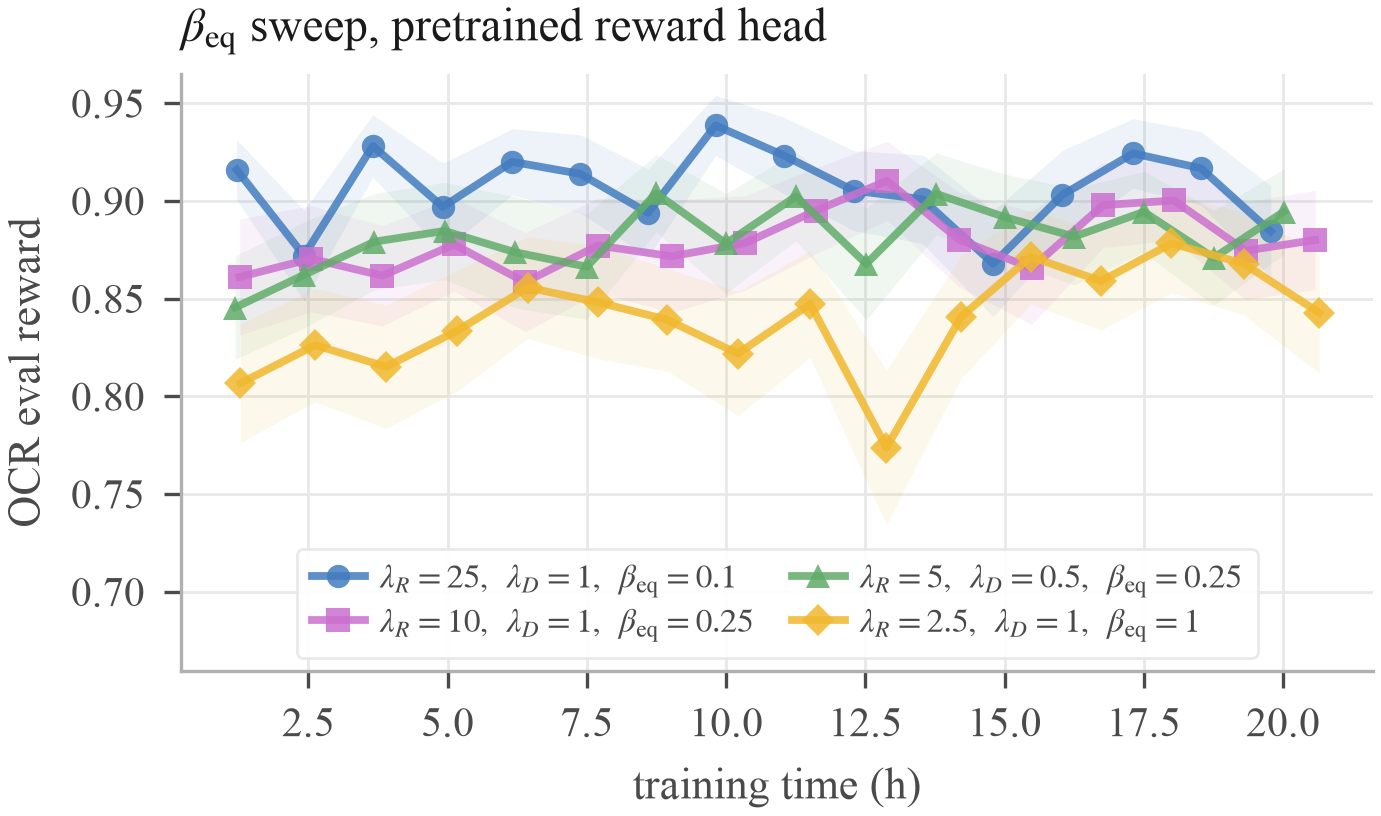}
    \caption{$\beta_{\mathrm{eq}}$ sweep, warmed-up reward head.}
    \label{fig:ablation-beta}
  \end{subfigure}%
  \begin{subfigure}{0.5\textwidth}
    \includegraphics[width=\linewidth]{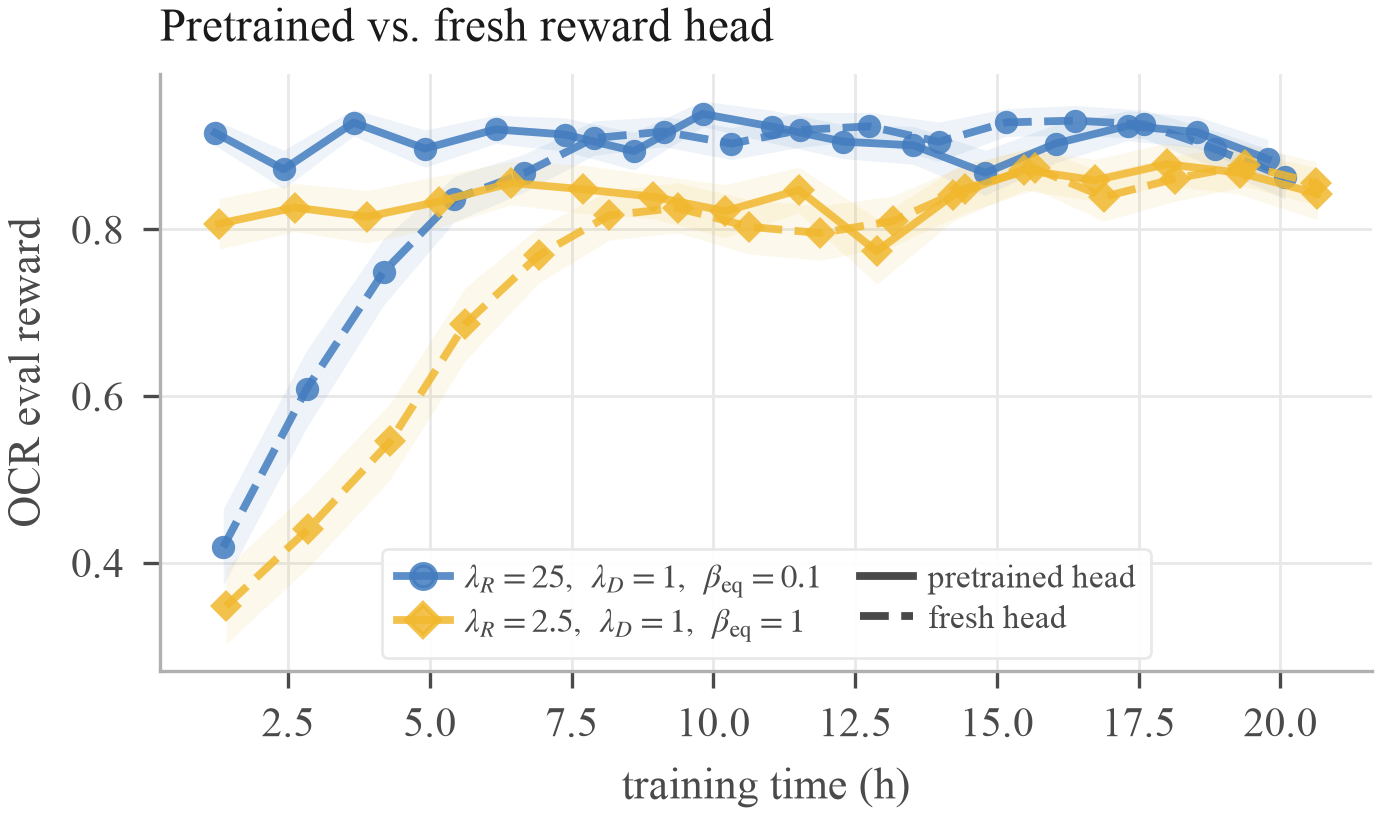}
    \caption{Warmed-up vs.\ fresh reward head.}
    \label{fig:ablation-head}
  \end{subfigure}
  \caption{{\textbf{Ablations on OCR.} Evaluation OCR score during training, (a) for several values of $\beta_{\mathrm{eq}}$ with a warmed-up reward head, and (b) for a warmed-up versus a freshly initialized reward head. 
  }}
  \label{fig:ablation}
\end{figure*}

\paragraph{Reward strength: Sensitivity to $\beta_{\mathrm{eq}}\propto \lambda_D/\lambda_R$.}
Since $\beta_{\mathrm{eq}} \propto C\,\lambda_D/\lambda_R$ with $C=2.5$ fixed, sweeping
$\beta_{\mathrm{eq}}$ is exactly sweeping the reward-to-reference weight ratio,
smaller $\beta_{\mathrm{eq}}$ meaning a more reward-weighted objective. OCR reward improves monotonically as $\beta_{\mathrm{eq}}$ falls (\cref{fig:ablation-beta}): over the
final four checkpoints, $0.907$ at $\beta_{\mathrm{eq}}=0.1$, $0.888$ at $0.25$ and
$0.862$ at $1$, against a $\pm 0.029$ confidence interval. The ratio being the
right parameter is confirmed by the two $\beta_{\mathrm{eq}}=0.25$ runs, which halve
the absolute weights from $(\lambda_D,\lambda_R)=(1,10)$ to $(0.5,5)$ and still
agree to $0.015$ across all sixteen checkpoints. {We see no collapse even at}
$\beta_{\mathrm{eq}}=0.1$, so the optimum may lie lower still.

\paragraph{{Warming up} the reward head buys speed, not quality.}
At two points of the sweep we initialize the reward head either
from {the warmed-up head used by default (\cref{sec:protocol})} or from scratch, holding everything else fixed
(\cref{fig:ablation-head}). A fresh head starts far behind: $0.497$ and $0.459$
below its warmed-up twin at the first checkpoint, as the generator 
{receives noisy scoring signals before the reward head is well-trained.} 
It recovers by step $3000$, some $8$
of the $\approx\!20$ wall-clock hours per run, and converges to within $0.003$ and
$0.004$ of the warmed-up runs, far inside the $\pm 0.029$ and $\pm 0.038$
intervals. {As this holds at both weight settings, the reward head warm-up only speeds up training, at a very small cost, and does not change the final result.}

\paragraph{Choice of data source for the reference pool.}
{For the same training prompts, we compare a \emph{base-model pool}, sampled from the base model, with a \emph{Nano Banana pool} generated through its API. The two differ well beyond their scores: base-model samples keep SD3.5-M's flat lighting, sparse layouts and misspelled words, whereas Nano Banana images are photographic, correctly lettered and set in fuller scenes. On OCR, the generator adopts the pool's style without losing the text (last two rows of \cref{fig:compare-verifiable}; more samples in \cref{fig:nanobananagen_tworow}): OCR stays at $0.92$ (\cref{tab:main}), the aesthetic score rises from $5.46$ to $5.62$, ImageReward, PickScore and HPS barely change, and CLIPScore drops from $28.75$ to $26.86$. GenEval, however, falls from $0.90$ to $0.73$, showing that the pool can also affect compositional accuracy. The two heads thus largely divide the work: the reward head governs what the reward measures, and the reference head governs everything the reward leaves unspecified, which is set by the {reference} pool.}

\section{Conclusion}
\label{sec:conclusion}

\begin{samepage}
{We presented \method{}, a simple method for adapting few-step diffusion generators using only external reward evaluations and reference samples. It turns KL-regularized reward maximization into two binary classification problems, solved by lightweight heads on a frozen diffusion backbone. With four steps, it reaches OCR and GenEval scores of $0.92$ and $0.90$, comparable to RL methods with 40--80 NFE, while improving all quality and preference scores over the base model, and gives the best HPSv3 on Z-Image-Turbo. The two heads divide the work: the reward head sets what the reward measures, and the reference pool largely sets the rest, such as style. \Cref{app:limitations} discusses limitations and future directions.}
\par
\end{samepage}

\clearpage
\label{page:lastmain}
\subsection*{AI use statement}

Generative AI tools were used in several parts of this work: to help structure the paper and draft and edit its text and equations, to search and summarize related literature, to inspect implementation and experiment records, and to suggest experimental controls and interpretations of results. The authors reviewed all of this material before using it; in particular, citations and literature summaries were checked against the original papers, and all reported numbers were checked against the experiment logs. The authors take responsibility for the final content of this work, including text, claims, and artifacts produced with the aid of generative AI.

\subsection*{Reproducibility statement}

\cref{sec:method} specifies the method, and \cref{alg:dualheads} summarizes one training iteration. \cref{sec:protocol} describes the base models, rewards, prompt sets, reference pools, and evaluation protocol, and \cref{app:implementation} gives the training cycle, head architecture, update schedule, and hyperparameters. The base models, prompt sets, and reward models we use are publicly available, and we will release the code, configuration files, and reference pools upon publication. 

\subsection*{Acknowledgments}
This work was supported by Hi!\ Paris (grant and fellowship), the ANR/France 2030 program (ANR-23-IACL-0005), ANR JCJC ``REEL-WORLD'', and a Google DeepMind academic gift. Computing resources were provided by GENCI through access to the IDRIS HPC facilities under allocations 2026-AD011014300R3, 2025-A0190616899, 2026-AD011015894R1, and by Google Gemini. The authors would especially like to thank Julie Mordacq and David Loiseaux for their help. 

\bibliography{shortstrings, references}
\bibliographystyle{iclr2027_conference}

\clearpage
\appendix

\begin{center}
    \Large\textbf{Appendix for \method{}}
\end{center}

This appendix is organized as follows:
\begin{itemize}
    \item \cref{app:limitations}: Limitations and future work.
    \item \cref{app:implementation}: Experimental details.
    \item \cref{app:results}: Additional results and visualizations.
\end{itemize}

\section{Limitations and future work}
\label{app:limitations}

{We see three main limitations.}
{\textbf{Idealized analysis.} Our analysis in \cref{sec:generator_objective} assumes optimal heads and a fixed reward head. In practice, the heads are only approximately optimal, and the reward head is refit on samples from the current generator, so the target moves during training. We also have no convergence guarantee for the alternating updates.}
{\textbf{Learned reward.} The generator is trained against the reward head, not the reward itself, so it inherits both the errors of the head and the flaws of the external reward. The reference head keeps samples close to the reference pool, but it cannot tell when the generator is exploiting the reward.}
{\textbf{Reference pool.} Fidelity is measured against the pool, so the generator also picks up properties that the reward ignores, such as style and composition (\cref{sec:exp-head}).}

{Several directions remain open. A convergence analysis of the alternating updates would put the method on firmer ground. The reward head also deserves a wider study, covering its architecture design, its backbone features analysis, the noise levels it sees, and so on. Richer reference pools, with several references per prompt or with real images, are another natural extension. Practically, \method{} could be applied during distillation, at higher resolutions, and to few-step video generators.}

\section{Experimental details}
\label{app:implementation}

This section describes one training cycle, the head architecture, the update schedule, the rewards, and the reference pools; \cref{sec:protocol} lists the base models and prompt sets.
Values in the text are those of the SD3.5-M OCR configuration; \cref{tab:hyperparameters} lists the configuration for each reward.
We use the EMA model for inference across the experiments.
The ablations use one NVIDIA H100 GPU and the main runs use four.

\subsection{Training cycle}
\label{app:training_cycle}

\paragraph{Scored rollout endpoints.}
For each prompt the generator runs the sampler \cref{eq:rollout} without gradient, with $N=4$ steps on the shifted time grid $(t_0,t_1,t_2,t_3)=(0,0.1,0.25,0.5)$.
The endpoints $x_1$ are decoded, scored with the reward $r$, and assigned the labels $y$ of \cref{eq:labels} with $C=2.5$.
The reward head is trained on $x_1$ re-noised to noise level $\sigma=1-t\sim\mathcal{U}[0.25,0.98]$, that is, $t\in[t_{\min},t_{\max}]=[0.02,0.75]$, with the image-level label $y$ repeated over the patch map.

\paragraph{Selected reference images.}
Each prompt is paired with at most one reference latent from the pool (\cref{sec:protocol}), repeated for every sample in its group.
The reference image is the positive example of the reference head and is re-noised over the same range $[t_{\min},t_{\max}]$.

\paragraph{Recomputed generator predictions.}
An endpoint $x_1$ is re-noised to one rollout grid timestep $t_i$, drawn once per group, giving $x_{t_i}=\mathcal{N}_{t_i}(x_1;\epsilon')$, and a single generator evaluation gives $\xhat_1=f_\theta(x_{t_i},c,t_i)$.
A no-gradient copy of $\xhat_1$, re-noised to $t\in[t_{\min},t_{\max}]$, is the negative example of the reference head.
The gradient copy, re-noised to an independent $t$, is scored by both heads to form the utility $s$, and the generator update is the pathwise gradient of $s$ through this single evaluation.

\subsection{Heads details}
\label{app:heads}

For SD3.5-M, for example, both heads read the hidden states of four blocks, with zero-based indices $\mathcal{T}=\{6,12,18,22\}$, of the frozen teacher transformer, reshaped to the $32\times32$ latent-patch grid at $512\times512$ resolution.
Each level $k\in\mathcal{T}$ has its own convolutional head: $3\times3$ convolutions of widths $1536\to640\to320$ with GroupNorm and FiLM modulation \citep{sauer2024fast} for conditions $(c,t)$, followed by a zero-initialized $1\times1$ output projection.
The level logits are summed,
\begin{equation}
  D_\phi(x_t,c,t)=\sum_{k\in\mathcal{T}} D_{\phi,k}\big(F_k(x_t,c,t),c,t\big),
  \label{eq:heads}
\end{equation}
and analogously for $R_\psi$, where $F_k$ is the feature map of block $k$.
The reward head is an independent copy of the reference head's architecture with its own optimizer.
Each head outputs a $32\times32$ map of patch logits; both BCE losses and the utility average over patches.

\subsection{Optimization and hyperparameters}
\label{app:hyperparameters}

\cref{tab:hyperparameters} lists the training configuration for each reward.
The generator keeps an EMA $\bar\theta_k=d_k\bar\theta_{k-1}+(1-d_k)\theta_k$; training rollouts use the raw weights.
The utility weights are set through $\beta_{\mathrm{eq}}$ (\cref{sec:generator_objective}), which \cref{sec:exp-head} sweeps, for instance with $(\lambda_D,\lambda_R)=(1,10)$ and $(0.5,5)$.

\begin{table}[!t]
\vspace{-20pt}
\centering
\small
\caption{Training configuration of the main runs for each reward.}
\label{tab:hyperparameters}
\resizebox{\linewidth}{!}{%
\begin{tabular}{@{}llcccc@{}}
\toprule
& & OCR & GenEval & Multi-reward & HPSv3 \\
\midrule
\multirow{2}{*}{Model} & Base model & SD3.5-M & SD3.5-M & SD3.5-M & Z-Image \\
& LoRA rank / $\alpha$ & \multicolumn{4}{c}{$32$ / $64$} \\
\midrule
\multirow{3}{*}{Sampling} & Resolution & \multicolumn{4}{c}{$512^2$} \\
& Sampler; steps; CFG & \multicolumn{4}{c}{LCM; $4$; $1$} \\
& Time grid & \multicolumn{4}{c}{$(0,0.1,0.25,0.5)$} \\
\midrule
\multirow{4}{*}{Rollouts} & Group size $K$ & \multicolumn{4}{c}{$24$} \\
& Prompt groups per epoch & \multicolumn{4}{c}{$48$} \\
& Label sharpness $C$ & $2.5$ & $2.5$ & $2.5$ / $0.5$ & $2.5$ \\
& Head noise level $\sigma=1-t$ & \multicolumn{4}{c}{$[0.25,0.98]$} \\
\midrule
Reference pool & Source & 5k OCR rows & $10$k GenEval rows & Pick-a-Pic images & Pick-a-Pic images \\
\midrule
\multirow{2}{*}{Heads} & Levels; widths & \multicolumn{3}{c}{$\{6,12,18,22\}$; $1536\to640\to320$} & $\{7,14,21,29\}$; $3840\to640\to320$ \\
& Parameters (both heads) & \multicolumn{3}{c}{$92.5$M} & $208.3$M \\
\midrule
\multirow{3}{*}{Optimization} & Generator & \multicolumn{4}{c}{AdamW, lr $2.5\times10^{-5}$, $(\beta_1,\beta_2)=(0.9,0.999)$, weight decay $10^{-4}$, grad.\ clip $1$} \\
& Heads & \multicolumn{4}{c}{AdamW, lr $10^{-4}$, $(\beta_1,\beta_2)=(0.9,0.999)$, weight decay $10^{-2}$, no grad.\ clip} \\
& EMA decay $d_k$ & \multicolumn{4}{c}{$\min\{(1+k)/(10+k),\,0.99\}$} \\
\midrule
Utility & $(\lambda_D,\lambda_R)$ & (1,10) & (1,50) & (1,25)& (1,25) \\
\midrule
Precision & & \multicolumn{4}{c}{BF16 autocast; FP32 parameters, optimizer states, and VAE} \\
\bottomrule
\end{tabular}}
\end{table}

\subsection{Rewards}
\label{app:rewards}

\paragraph{OCR.}
Text is recognized with PaddleOCR~\citep{cui2025paddleocr}.
The target string $w$ is the first double-quoted span of the prompt; prompts without a non-empty target are excluded before scoring.
Recognized segments with positive confidence are concatenated, in the order returned by the OCR engine, into a string $\hat w$.
Both strings are lowercased and ASCII spaces are removed; other whitespace, punctuation, and Unicode forms are left unchanged.
The reward is
\begin{equation}
 r_{\mathrm{OCR}}(\hat w,w)=
 \begin{cases}
 1, & \text{if $w$ is a substring of $\hat w$},\\
 \max\{0,\,1-d_{\mathrm{Lev}}(\hat w,w)/|w|\}, & \text{otherwise},
 \end{cases}
 \label{eq:ocr-reward}
\end{equation}
where $d_{\mathrm{Lev}}$ is the Levenshtein distance and $|w|$ the target length; recognition failures receive zero.
The reward therefore tolerates extra text and ignores case and spacing.

\paragraph{GenEval.}
The generator is trained on the GenEval training prompts from Flow-GRPO, with a reference pool of 10{,}000 rows sampled from the GenEval training set (\cref{sec:protocol}).
{We score images with the GenEval evaluator~\citep{ghosh2023geneval}, as implemented in the Flow-GRPO reward server~\citep{liu2025flow}: Mask2Former~\citep{cheng2022masked} (Swin-S, trained on COCO) detects objects with confidence above $0.3$ ($0.9$ for counting), CLIP ViT-L/14~\citep{radford2021learning} classifies their colors, and relative positions are read from box centroids.
A requested object with count $n$ and $\hat n$ detections earns the credit $1-|n-\hat n|/n$ and, only if $\hat n=n$, a further credit for its color or position.
The \emph{dense} training reward is the mean of these credits, while the \emph{strict} score is $1$ only if every requirement is met exactly.
\cref{tab:main,tab:geneval} report the official GenEval accuracy, averaged over its six task categories.}

\paragraph{Multi-reward (PickScore + CLIPScore + HPSv2.1).}
The generator is trained on Pick-a-Pic prompts, with the existing Pick-a-Pic images as the reference pool (\cref{sec:protocol}).
{The three scores are PickScore~\citep{kirstain2023pick} (the \texttt{PickScore\_v1} checkpoint, on a CLIP ViT-H/14 backbone), CLIPScore~\citep{hessel2021clipscore} with OpenAI CLIP ViT-L/14~\citep{radford2021learning}, and HPSv2.1~\citep{wu2023human} (on a CLIP ViT-H/14 backbone).
Following DiffusionNFT~\citep{zheng2025diffusionnft}, the training reward is the sum of the three scores with equal weights, and the labels of \cref{eq:labels} are computed from this sum.}

\paragraph{HPSv3 (Z-Image).}
{The generator is initialized from Z-Image-Turbo~\citep{jiang2026distribution}, the few-step distilled version of Z-Image~\citep{zimage2025}, and fine-tuned with the same LoRA configuration as SD3.5-M (\cref{sec:protocol}).
The reward is HPSv3~\citep{ma2025hpsv3}, with the released checkpoint built on Qwen2-VL.
As in the multi-reward setting, we train on Pick-a-Pic prompts with the Pick-a-Pic images as the reference pool, and evaluate on DrawBench prompts.
The reward weight is $\lambda_R=25$, and \cref{tab:zimage} reports EMA weights after 120 training iterations, sampled with two and four steps; all other settings follow \cref{tab:hyperparameters}.}

\section{Additional results}
\label{app:results}

\subsection{Multi-reward training}
\label{app:multi_reward}

{We train SD3.5-M jointly on PickScore, CLIPScore and HPSv2.1 with Pick-a-Pic prompts, using the Pick-a-Pic images as the reference pool (\cref{sec:protocol}), and evaluate on DrawBench prompts; \cref{tab:multi_reward} reports the results.}

\begin{table}[!t]
\vspace{-20pt}
\centering
\footnotesize
\setlength{\tabcolsep}{3pt}
\caption{Comparison on multi-reward training with PickScore, CLIPScore and HPSv2.1 as training rewards ($^{\ast}$). Image quality and preference scores are computed on \texttt{DrawBench}~\citep{saharia2022photorealistic} prompts. Rows marked $^{\dagger}$ are quoted from Table~1 of MeanFlowNFT~\citep{huang2026meanflownft}, and the row marked $^{\ddagger}$ from \citet{choi2026rethinking}; the other rows are our evaluations. Abbreviations: \texttt{Aes.} (Aesthetics), \texttt{ImgRwd} (ImageReward), \texttt{Pick} (PickScore), \texttt{CLIP} (CLIPScore), \texttt{HPS} (HPSv2.1).}
\label{tab:multi_reward}
\resizebox{\linewidth}{!}{%
\begin{tabular}{@{}clcccccc@{}}
\toprule
& \textbf{Method} & \texttt{NFE} & \texttt{Aes.}$\uparrow$ & \texttt{ImgRwd}$\uparrow$ & \texttt{Pick}$^{\ast}\uparrow$ & \texttt{CLIP}$^{\ast}\uparrow$ & \texttt{HPS}$^{\ast}\uparrow$ \\
\midrule
& SD3.5-M (base) & 80 & 5.39 & 0.87 & 22.34 & 27.99 & 0.279 \\
\cmidrule(lr){1-8}
& SD3.5-M w/o CFG$^{\dagger}$ & 80 & 5.15 & $-$0.48 & 20.66 & 23.91 & 0.211 \\
& SD3.5-M w/ CFG$^{\dagger}$ & 80 & 5.38 & 0.93 & 22.46 & 28.80 & 0.282 \\
& Flow-GRPO~\citep{liu2025flow} w/ CFG$^{\dagger}$ & 80 & 5.28 & 1.02 & 22.48 & 29.12 & 0.266 \\
& DiffusionNFT~\citep{zheng2025diffusionnft} w/o CFG$^{\dagger}$ & 40 & 5.96 & 1.41 & 23.64 & 28.89 & 0.324 \\
& \citet{choi2026rethinking}$^{\ddagger}$ & 40 & 6.06 & 1.45 & 23.68 & 29.60 & 0.325 \\
\multirow{-6}{*}{\rotatebox[origin=c]{90}{\scriptsize Multi-step}} & CRD~\citep{zhu2026diffusion} & 40 & 5.70 & 1.35 & 23.27 & 29.63 & 0.321 \\
\cmidrule(lr){1-8}
& DMD~\citep{yin2024one}$^{\dagger}$ & 4 & 5.51 & 0.92 & 22.28 & 28.41 & 0.287 \\
& RTDMD~\citep{huang2026reinforcing}$^{\dagger}$ & 4 & 6.13 & 1.23 & 23.28 & 27.75 & 0.327 \\
& $R_{\mathrm{dm}}$~\citep{fan2026r}$^{\dagger}$ & 4 & 5.75 & 0.72 & 22.07 & 27.20 & 0.276 \\
& DMD + DiffusionNFT$^{\dagger}$ & 4 & 5.38 & 0.72 & 21.96 & 28.43 & 0.271 \\
& CDM + DiffusionNFT$^{\dagger}$ & 4 & 4.83 & 0.15 & 21.38 & 27.45 & 0.214 \\
& AnyFlow + DiffusionNFT$^{\dagger}$ & 4 & 5.95 & 1.24 & 23.09 & 29.15 & 0.292 \\
& MeanFlowNFT~\citep{huang2026meanflownft}$^{\dagger}$ & 4 & 5.93 & 1.45 & 23.50 & 29.67 & 0.327 \\
\rowcolor{gray!15} \multirow{-8}{*}{\rotatebox[origin=c]{90}{\scriptsize Few-step}} & \textbf{\method{}}  & 4 & 5.74 & 1.22 & 22.76 & 28.18 & 0.320 \\
\bottomrule
\end{tabular}%
}
\end{table}

\subsection{Per-category GenEval results}
\label{app:geneval}

{\cref{tab:geneval} breaks down the \texttt{GenEval} scores of \cref{tab:main} by category.}

\begin{table*}[!ht]
\vspace{-20pt}
    \centering
    \footnotesize
    \setlength{\tabcolsep}{4pt}
    \caption{\textbf{\texttt{GenEval} results by category.} Baseline numbers are taken from Flow-GRPO~\citep{liu2025flow} or the original papers, and reimplemented where unavailable. Among the RL-tuned models, the best score per column is in \textbf{bold} and the second best is \underline{underlined}; the shaded row is our method. Obj.: object; Attr. Bind.: attribute binding.}
\resizebox{\textwidth}{!}{%
    \begin{tabular}{@{}lccccccc@{}}
    \toprule
    \textbf{Model} & Single Obj. & Two Obj. & Counting & Colors & Position & Attr. Bind. & Overall \\
    \midrule
    \multicolumn{8}{@{}l}{\textit{Diffusion and flow-matching models}} \\
    \cmidrule(r){1-1}
    SD-XL~\citep{podell2023sdxl} & 0.98 & 0.74 & 0.39 & 0.85 & 0.15 & 0.23 & 0.55 \\
    FLUX.1 Dev~\citep{flux2024} & 0.98 & 0.81 & 0.74 & 0.79 & 0.22 & 0.45 & 0.66 \\
    SD3.5-M~\citep{esser2024scaling} & 0.98 & 0.78 & 0.50 & 0.81 & 0.24 & 0.52 & 0.63 \\
    TDM-SD3.5-M~\citep{luo2025learning} (4 NFE) & 0.99 & 0.77 & 0.49 & 0.79 & 0.23 & 0.44 & 0.61 \\
    \addlinespace
    \multicolumn{8}{@{}l}{\textit{Multi-step RL (SD3.5-M backbone)}} \\
    \cmidrule(r){1-1}
    Flow-GRPO~\citep{liu2025flow} & \textbf{1.00} & \underline{0.99} & 0.95 & \textbf{0.92} & \textbf{0.99} & 0.86 & \underline{0.95} \\
    AWM~\citep{xue2025advantage} (w/o CFG) & \textbf{1.00} & 0.98 & 0.95 & 0.77 & 0.78 & 0.65 & 0.91 \\
    DiffusionNFT~\citep{zheng2025diffusionnft} & \textbf{1.00} & \underline{0.99} & \underline{0.96} & 0.82 & 0.78 & 0.73 & 0.92 \\
    CRD~\citep{zhu2026diffusion} & \textbf{1.00} & 0.98 & 0.92 & 0.88 & 0.90 & 0.73 & 0.93 \\
    DMSampler~\citep{liudistillation} & \textbf{1.00} & \textbf{1.00} & \textbf{0.98} & \underline{0.90} & \textbf{0.99} & 0.86 & \textbf{0.96} \\
    \addlinespace
    \multicolumn{8}{@{}l}{\textit{Few-step RL (SD3.5-M backbone)}} \\
    \cmidrule(r){1-1}
    TDM-R1~\citep{luo2026tdm} & \textbf{1.00} & 0.96 & 0.88 & 0.85 & \underline{0.93} & \underline{0.91} & 0.92 \\
    RTDMD~\citep{huang2026reinforcing} & \textbf{1.00} & 0.98 & 0.95 & \underline{0.90} & 0.87 & \textbf{0.94} & 0.94 \\
    \rowcolor{gray!15} \method{} & \textbf{1.00} & 0.98 & 0.91 & 0.85 & 0.90 & 0.76 & 0.90 \\
    \bottomrule
    \end{tabular}%
}
\label{tab:geneval}
\end{table*}

\subsection{Qualitative results}
\label{app:qualitative}

\begin{figure}[ht]
\centering
\includegraphics[width=\linewidth]{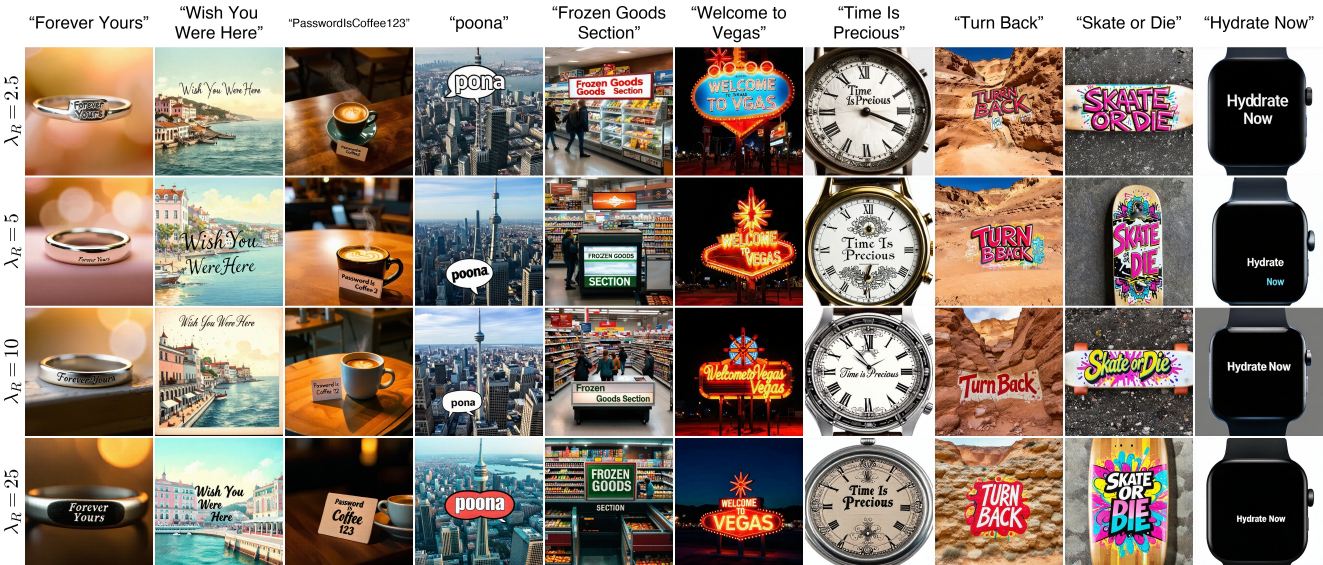}
\caption{Visualization of samples from the ablation in~\cref{fig:ablation-beta}.}
\label{fig:qual-ocr}
\end{figure}

\paragraph{Effect of the reward weight.}
\cref{fig:qual-ocr} shows samples from the reward-weight sweep of \cref{fig:ablation-beta}, with $\lambda_D=1$, $C=2.5$, and $\lambda_R\in\{2.5,5,10,25\}$ from top to bottom; each column is one OCR test prompt.
At $\lambda_R=2.5$ the text is often small or misspelled, for example ``pona'' and ``Hyddrate Now''.
As $\lambda_R$ grows, the text becomes larger and easier to read and these errors disappear, in line with the higher held-out OCR reward in \cref{fig:ablation-beta}.
At $\lambda_R=25$ the text also tends to dominate the image, often as a bold sign or sticker, while the layout of each scene stays similar across rows.

\begin{figure}[ht]
\centering
\begin{subfigure}{\linewidth}
\centering
\includegraphics[width=\linewidth]{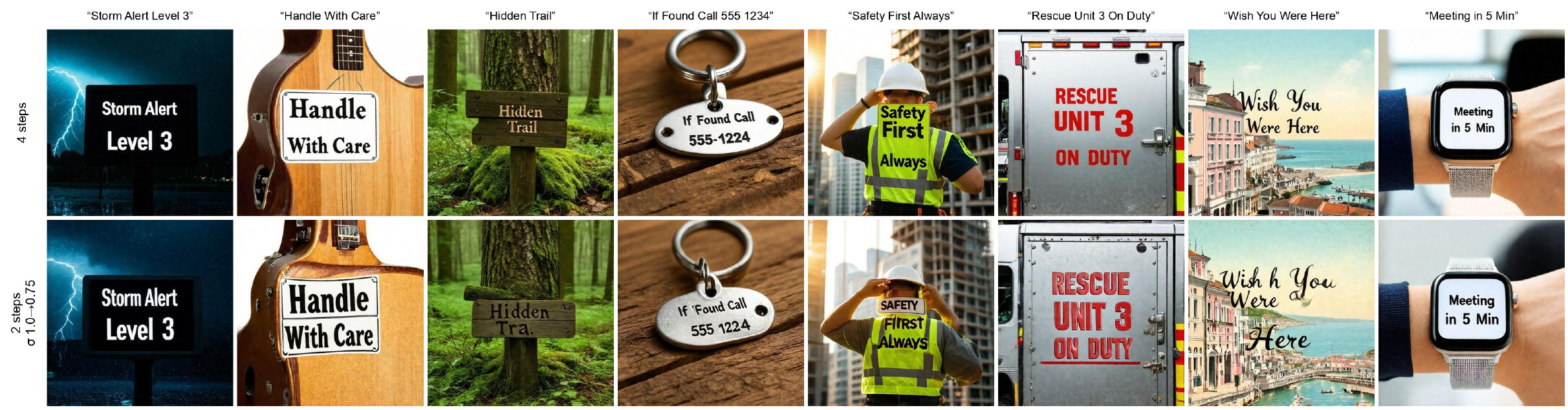}
\label{fig:2step-ocr}
\end{subfigure}
\par\vspace{-6pt}
\begin{subfigure}{\linewidth}
\centering
\includegraphics[width=\linewidth]{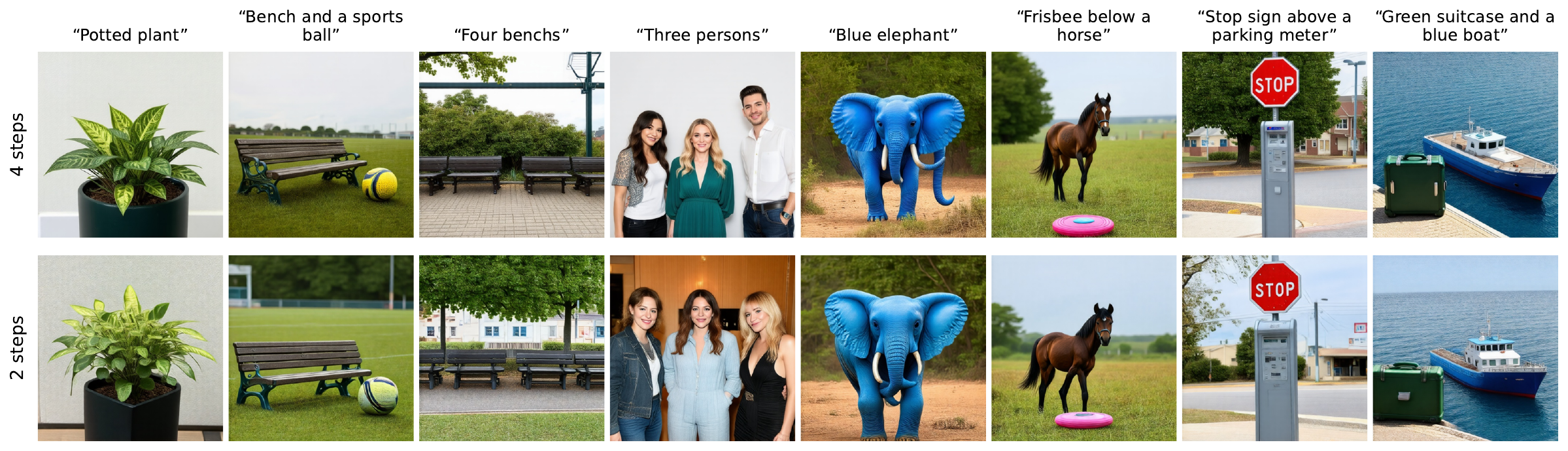}
\label{fig:2step-geneval}
\end{subfigure}
\caption{{Visual comparison of four-step and two-step generation from the four-step generators trained with the OCR reward (top) and the GenEval reward (bottom).}}
\label{fig:2step}
\end{figure}

\paragraph{Two-step sampling.}
{The generators are trained with four sampling steps, but they can also be sampled with two. \cref{fig:2step} shows four-step and two-step samples on the same prompts, for the OCR generator (top) and the GenEval generator (bottom). With two steps, the generator still reaches a reward close to that of four steps (\cref{tab:main}).}

\begin{figure}[t]
\centering
\includegraphics[width=\linewidth]{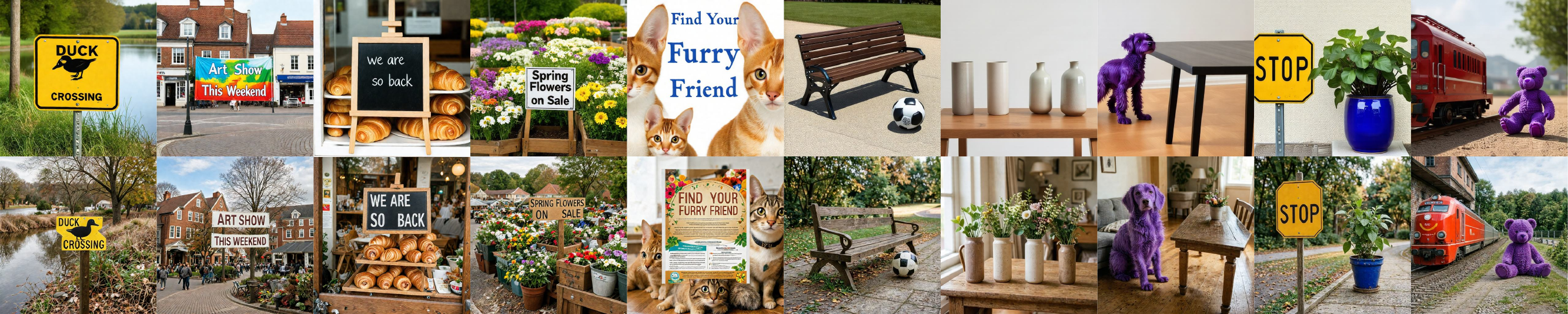}
\caption{\textbf{Effect of the reference pool.}
Same prompts with base-model (top) and Nano Banana (bottom) references: the style changes, the content does not.}
\label{fig:nanobananagen_tworow}
\end{figure}

\paragraph{Effect of the reference pool.}
\cref{fig:nanobananagen_tworow} shows that the reference pool sets the style of the generator but not the content it renders.

\begin{figure}[t]
\centering
\includegraphics[width=\linewidth]{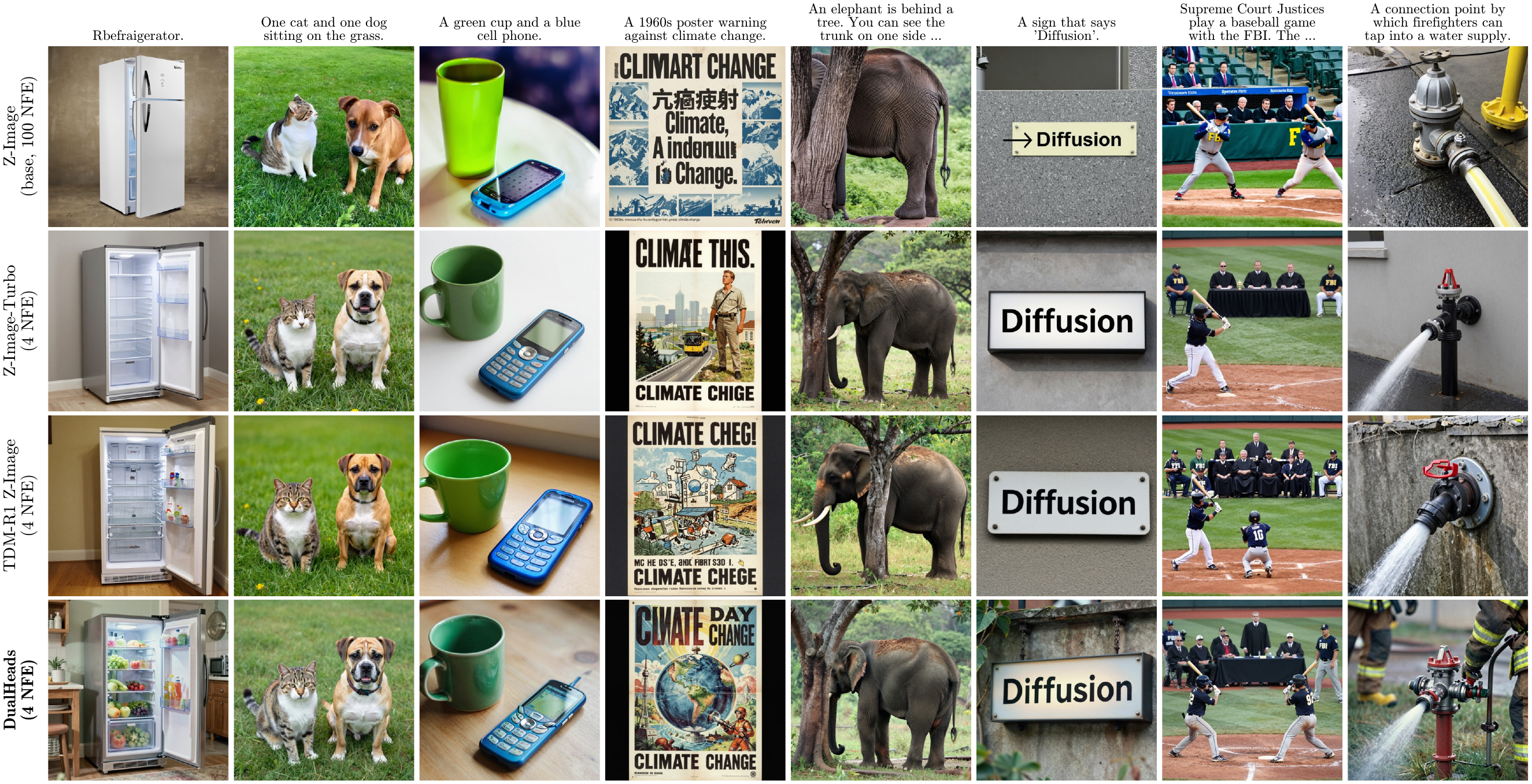}
\caption{{Qualitative comparison for Z-Image-based models.} 
}
\label{fig:compare-model-based}
\end{figure}

\paragraph{Comparison of Z-Image-based models.} \cref{fig:compare-model-based} shows samples from the Z-Image base model alongside its distilled and RL-trained variants.

\begin{figure}[ht]
\centering
\vspace{-20pt}
\includegraphics[width=0.85\linewidth]{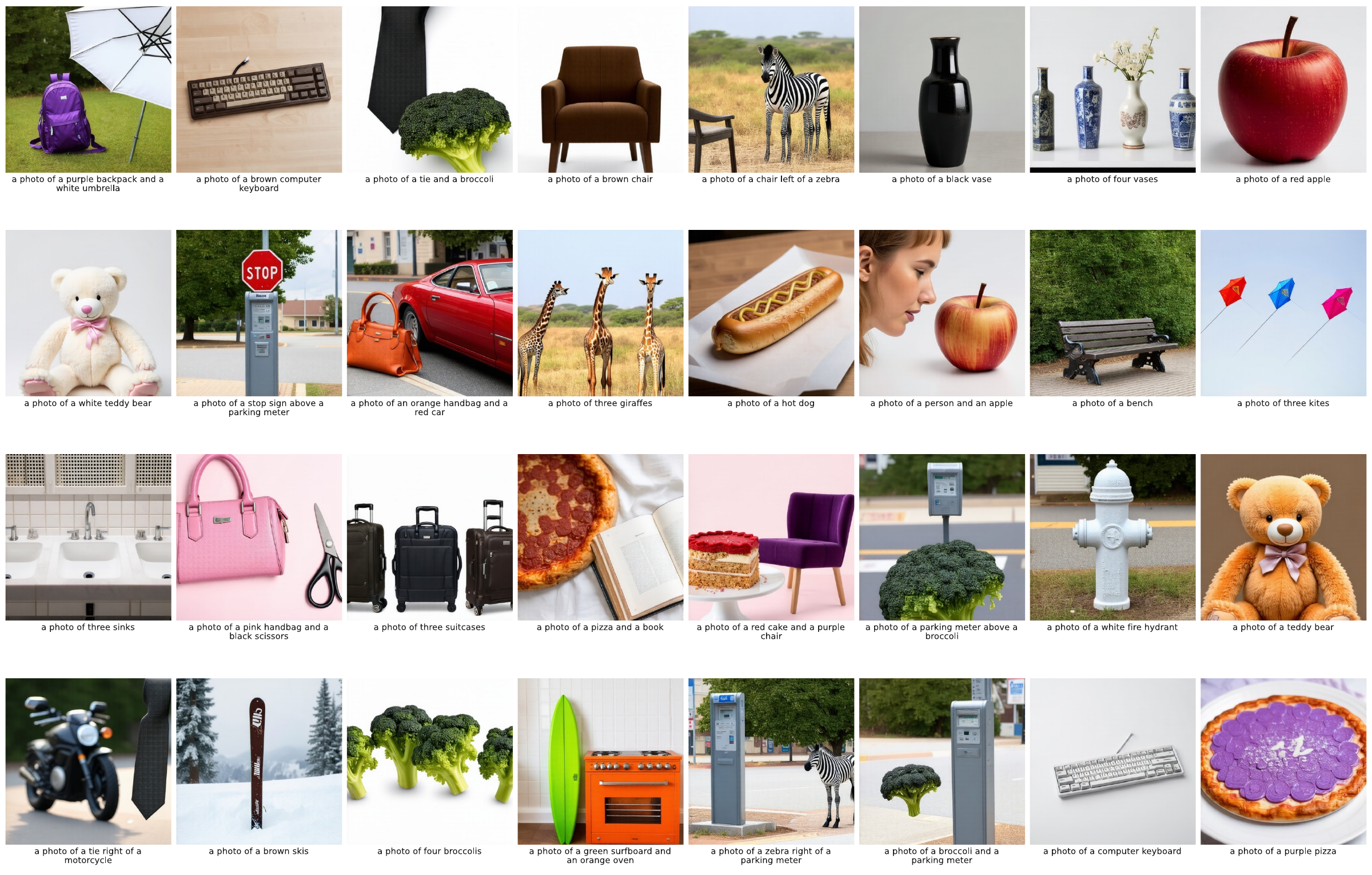}
\caption{Uncurated samples on GenEval test prompts.}
\label{fig:qual-geneval}
\end{figure}

\begin{figure}[ht]
\centering
\includegraphics[width=0.85\linewidth]{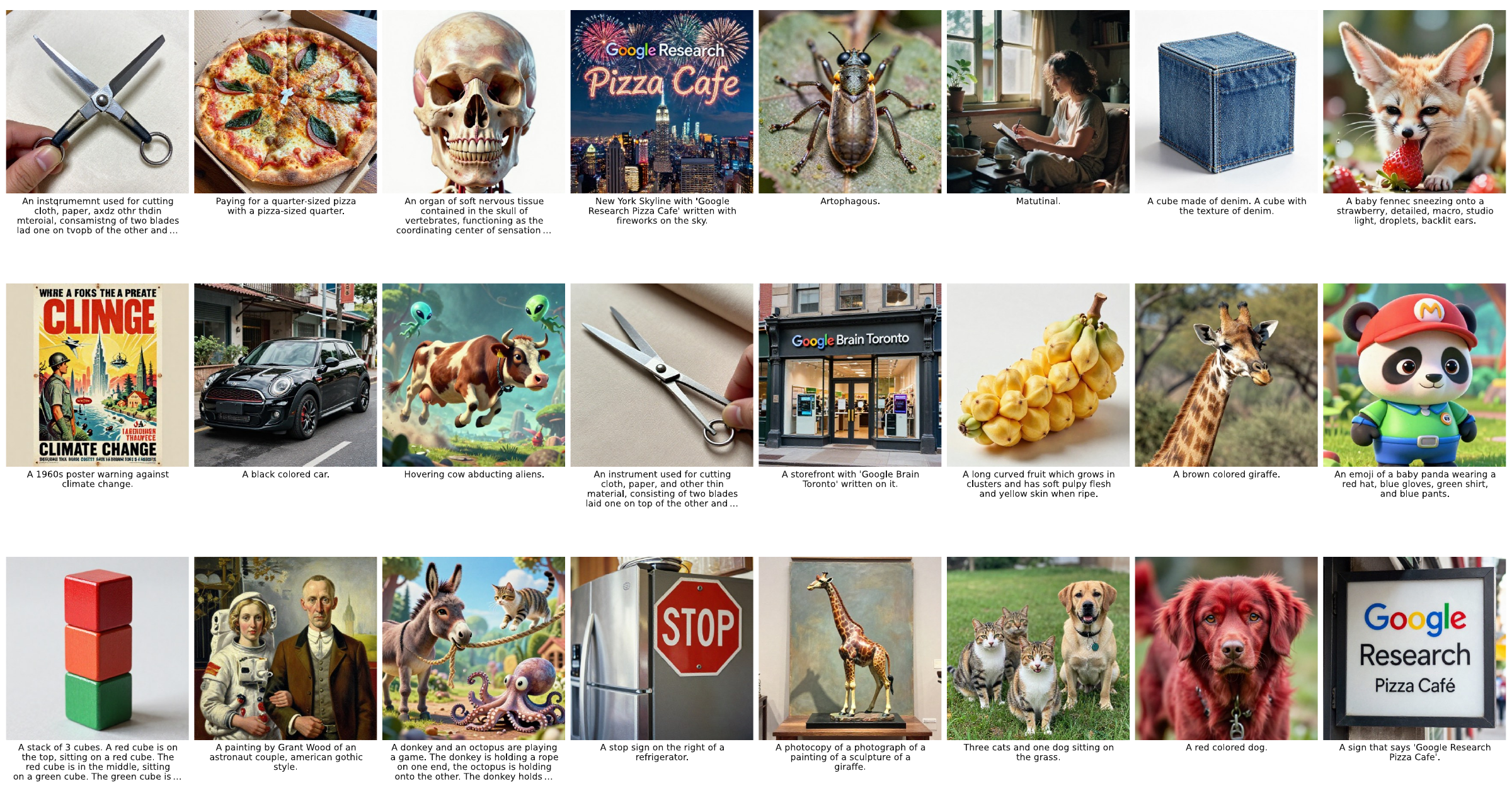}
\caption{Uncurated Z-Image samples after HPSv3 training.}
\label{fig:qual-zimage}
\end{figure}

\begin{figure}[ht]
\centering
\includegraphics[width=0.85\linewidth]{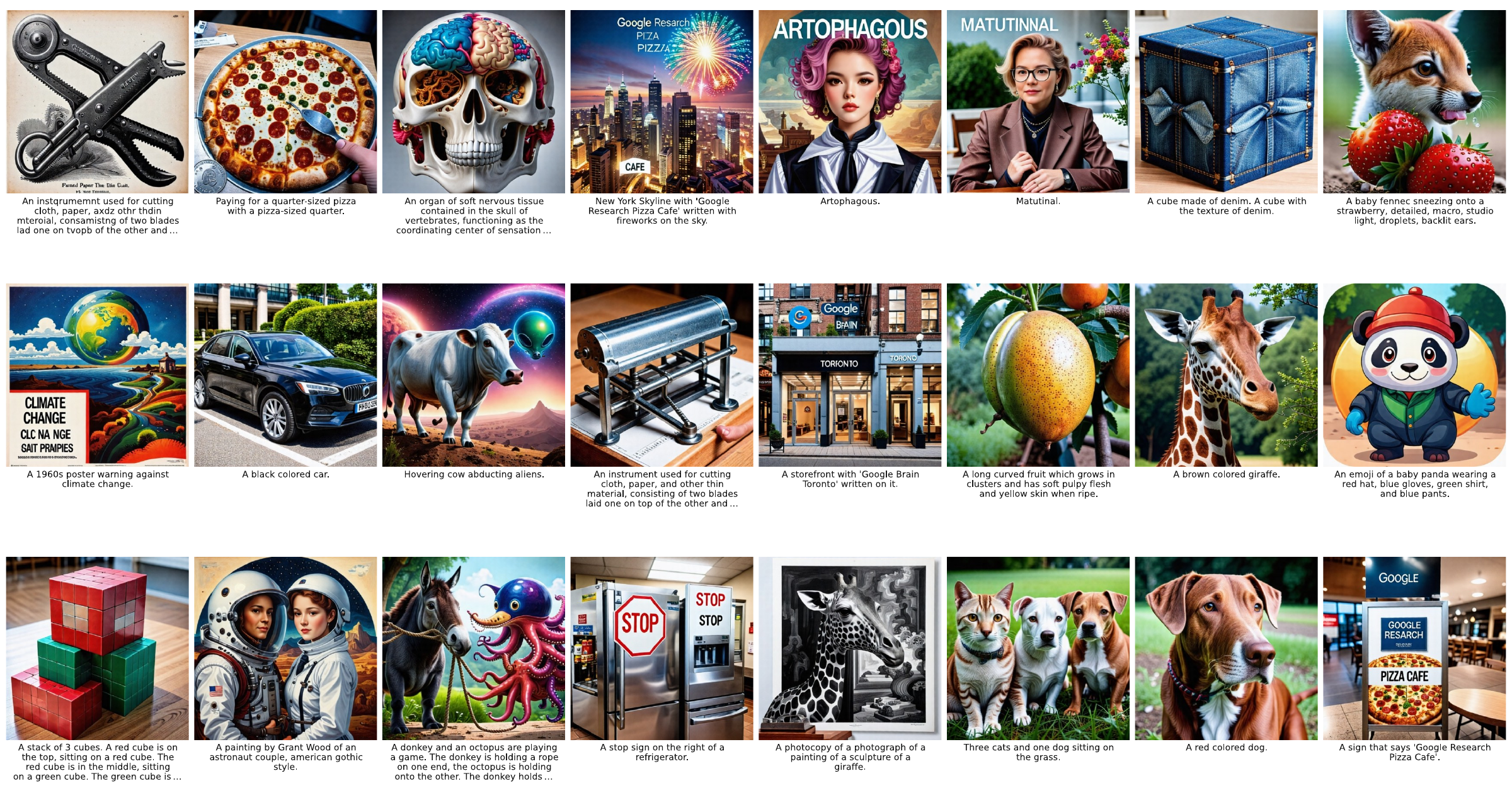}
\caption{Uncurated samples on DrawBench prompts after multi-reward training (PickScore + CLIPScore + HPSv2.1).}
\label{fig:qual-multi}
\end{figure}

\paragraph{Uncurated samples.}
{\cref{fig:qual-geneval,fig:qual-multi,fig:qual-zimage} show \method{} samples only, taken from the evaluation runs.
Prompts are drawn uniformly without replacement.}

\clearpage

\subsection{Samples from the reference dataset}
\label{app:ref_data}

We consider two sources of reference data for each task: images generated with Nano Banana, and teacher-simulated images sampled from the base model SD3.5-M at $512 \times 512$ with 20 steps and guidance scale 4.5.
For the teacher, we draw eight samples for each of the 5,000 OCR and 10,000 GenEval training prompts, and 24 more for the 1,931 OCR prompts whose first eight contain no good score.
For both sources, each prompt's highest-reward sample serves as its reference if that reward is at least 0.6 (OCR) or 0.5 (GenEval).

\Cref{tab:ref_reward_stats} shows that the kept references score far above these thresholds (mean $r$ of 0.90--0.98), because Nano Banana's rewards are mostly all-or-nothing and the teacher keeps the best of several draws.
The two sources thus differ mainly in coverage: the teacher provides a reference for 93--97\% of the prompts and Nano Banana for 62--63\%.

\begin{figure}[ht]
\centering
\includegraphics[width=0.99\linewidth]{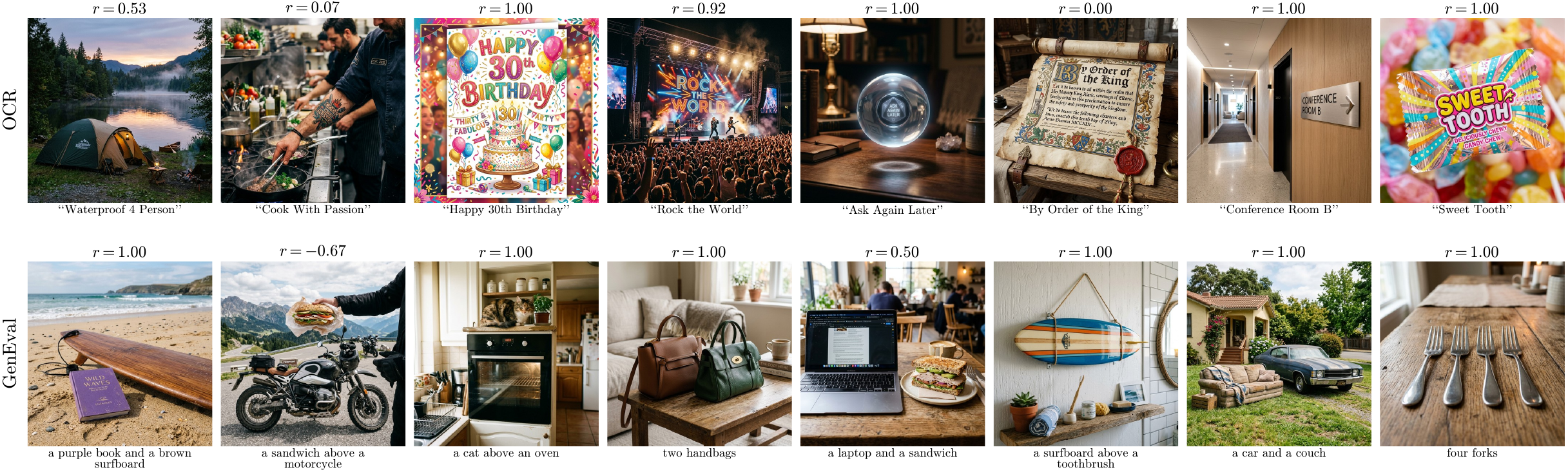}
\caption{\textbf{Random Nano Banana samples.}
Images for the OCR (top) and GenEval (bottom) training prompts, drawn at random.
Above each image is its reward $r$, the OCR reward or the dense GenEval score; below is the target text (OCR) or the prompt (GenEval).}
\label{fig:ref_random}
\end{figure}

\begin{figure}[ht]
\centering
\vspace{-25pt}
\includegraphics[width=0.99\linewidth]{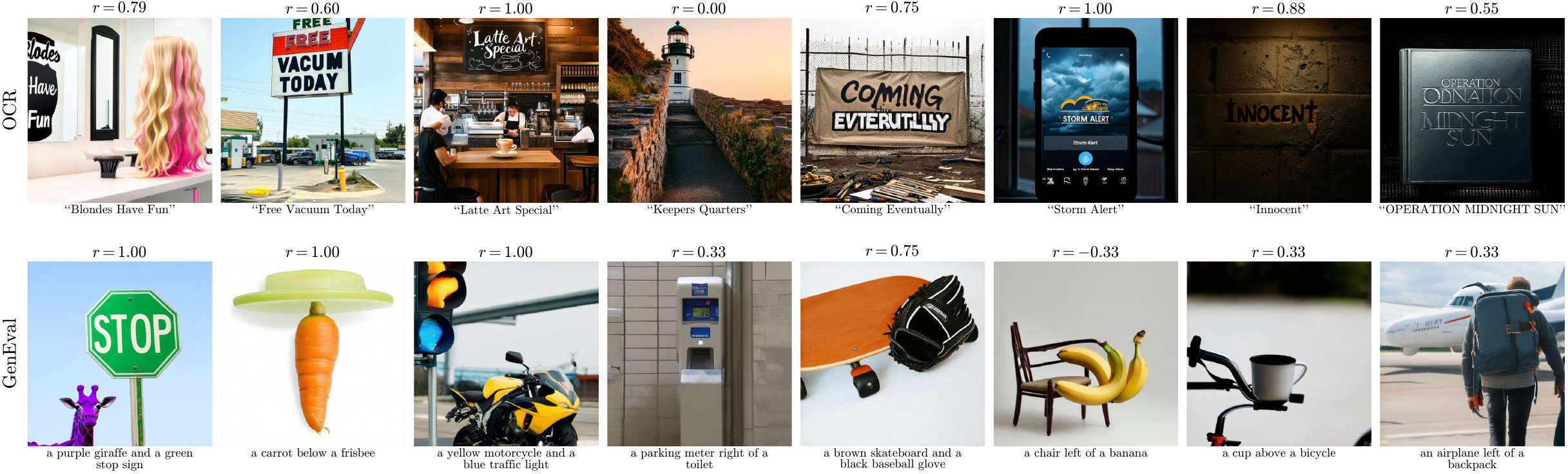}
\caption{\textbf{Random teacher-simulated samples.}
SD3.5-M samples for the OCR (top) and GenEval (bottom) training prompts, drawn at random; annotations as in \cref{fig:ref_random}.}
\label{fig:ref_teacher_random}
\end{figure}

\begin{figure}[ht]
\centering
\includegraphics[width=0.99\linewidth]{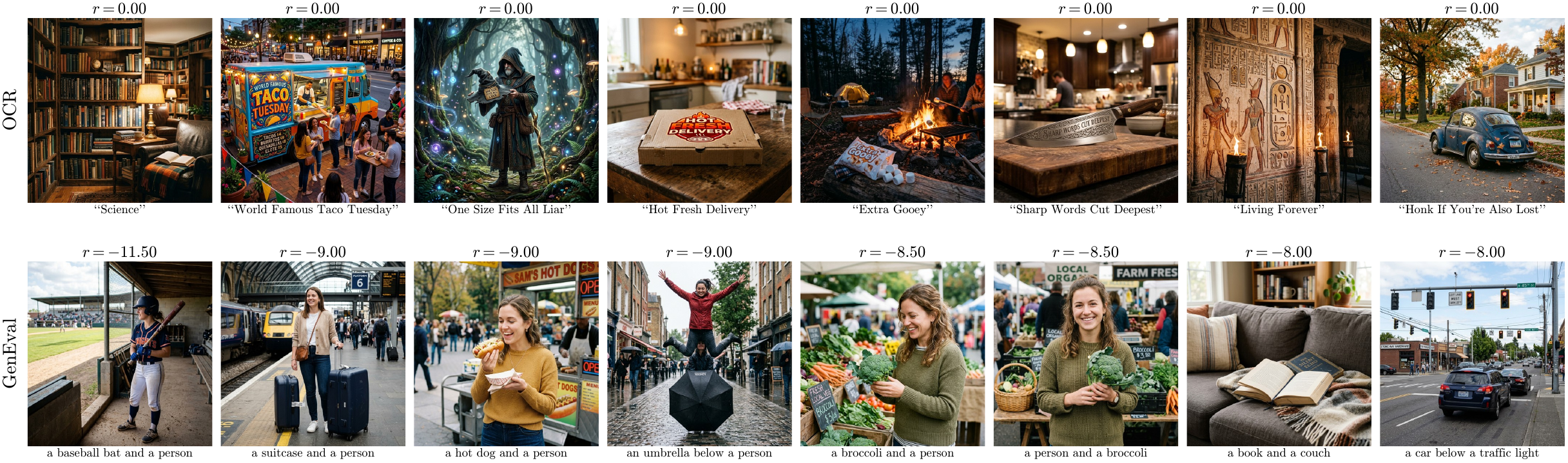}
\caption{\textbf{Lowest-reward Nano Banana samples.}
OCR (top): eight images drawn at random among those with $r = 0$, the minimum, which about 29\% of images reach.
GenEval (bottom): the eight images with the lowest dense GenEval score.}
\label{fig:ref_lowest}
\end{figure}

\begin{figure}[ht]
\centering
\includegraphics[width=0.99\linewidth]{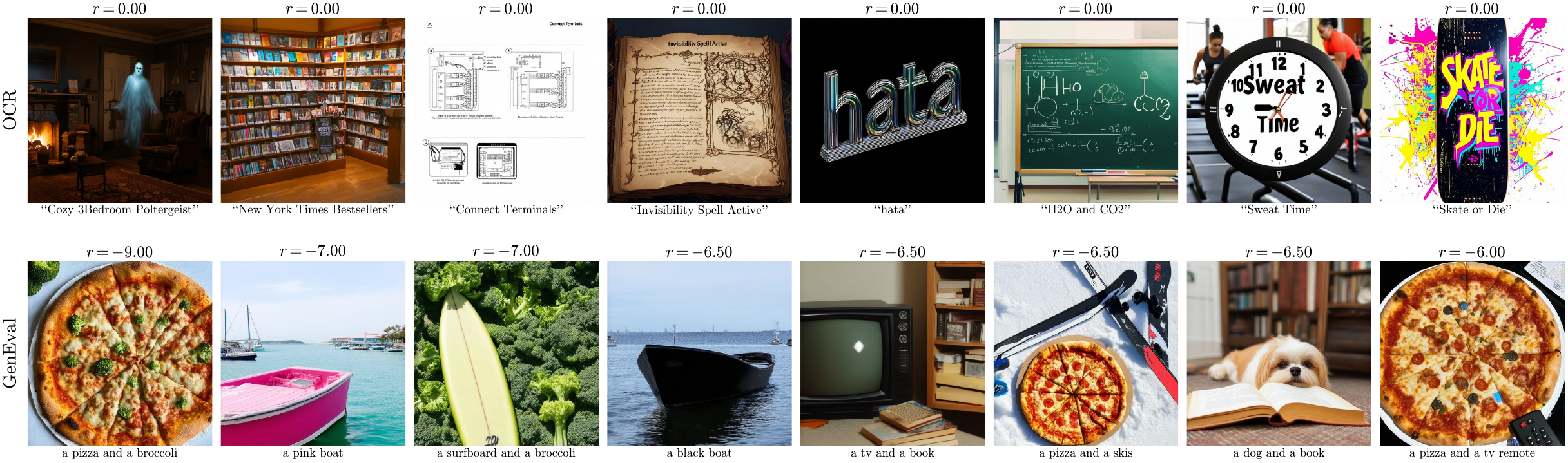}
\caption{\textbf{Lowest-reward teacher-simulated samples.}
OCR (top): eight samples drawn at random among those with $r = 0$, which about 20\% of samples reach.
GenEval (bottom): the eight lowest dense GenEval scores, at most one sample per prompt.}
\label{fig:ref_teacher_lowest}
\end{figure}

\Cref{fig:ref_random,fig:ref_teacher_random} show random samples from the two sources, annotated with their reward $r$.
\Cref{fig:ref_lowest,fig:ref_teacher_lowest} show the lowest-reward samples, many of which are correct but misjudged by the reward: stylized text that the OCR model fails to read, and cluttered scenes (broccoli on a pizza, books on a shelf) whose extra detections drive the dense GenEval score far below zero.
The teacher's lowest-reward samples also include genuine failures, such as missing or illegible text.

\begin{table}[t]
\centering
\vspace{-20pt}
\caption{\textbf{Reward statistics of the reference data.}
All samples: every generated image (for the teacher, the first eight draws per prompt).
References: each prompt's highest-reward sample with $r \ge 0.6$ (OCR) or $r \ge 0.5$ (GenEval); Kept is the share of prompts that have one.}
\label{tab:ref_reward_stats}
\small
\begin{tabular}{lrrrrrrrr}
\toprule
& & & \multicolumn{3}{c}{All samples} & \multicolumn{3}{c}{References} \\
\cmidrule(lr){4-6}\cmidrule(lr){7-9}
Source & Prompts & Samples & Mean $r$ & $r = 1$ & $r \le 0$ & Kept & Mean $r$ & $r = 1$ \\
\midrule
\multicolumn{9}{l}{\textit{OCR}} \\
Nano Banana & 19,621 & 19,625 & 0.63 & 51.9\% & 28.8\% & 61.6\% & 0.97 & 84.3\% \\
Teacher & 5,000 & 40,000 & 0.57 & 21.2\% & 20.5\% & 96.8\% & 0.98 & 81.1\% \\
\midrule
\multicolumn{9}{l}{\textit{GenEval}} \\
Nano Banana & 33,199 & 50,000 & 0.47 & 58.0\% & 20.2\% & 63.3\% & 0.93 & 81.9\% \\
Teacher & 10,000 & 80,000 & 0.48 & 28.8\% & 18.5\% & 92.7\% & 0.90 & 71.2\% \\
\bottomrule
\end{tabular}
\end{table}

\subsection{Ablations of the heads and of the generator loss type}
\begin{figure}[ht]
\centering
\includegraphics[width=\linewidth]{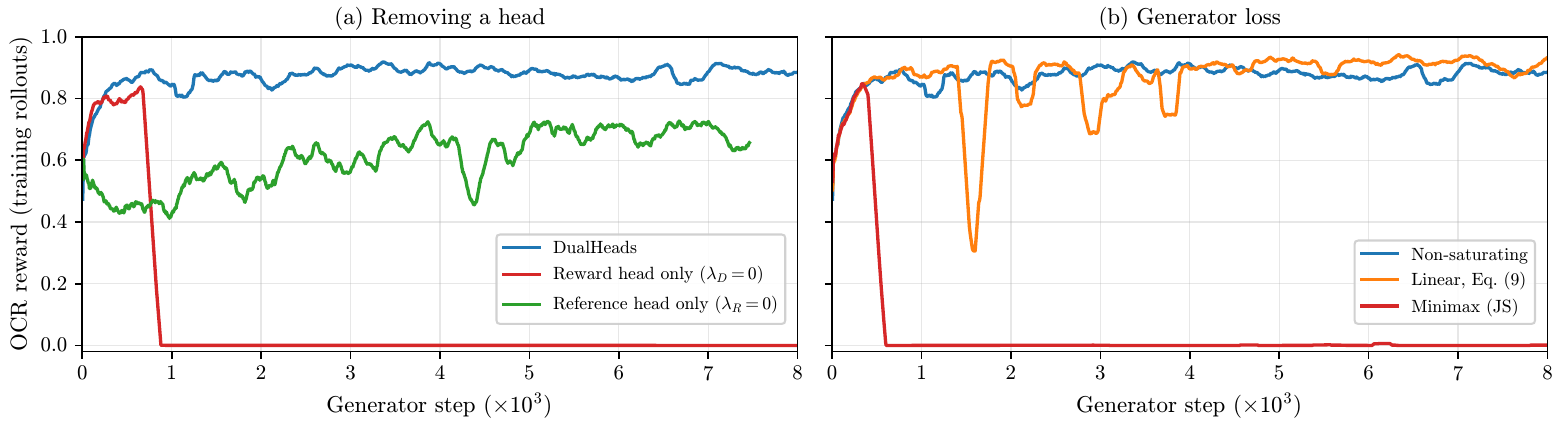}
\caption{\textbf{Ablations of the heads and of the generator loss type on OCR.}}
\label{fig:head_loss_ablation}
\end{figure}

\Cref{fig:head_loss_ablation} plots the OCR reward of the rollout samples during training.
Without the reference head ($\lambda_D = 0$), the reward first rises as with the full method, then drops to zero after about 900 steps: the training collapses.
Without the reward head ($\lambda_R = 0$), samples stay natural, but the reference head alone does not work as well. The training reward rises only slightly, likely due to the high-quality data from the reference pool.
In addition, we show results with three classic generator training loss types.

\clearpage

\end{document}